\RequirePackage[dvipsnames]{xcolor}
\documentclass{article}
\PassOptionsToPackage{numbers, compress}{natbib}
\usepackage[final]{neurips_data_2022}
\usepackage[utf8]{inputenc}

\usepackage{amssymb,graphicx}
\usepackage{subcaption}
\usepackage{amsmath}
\usepackage[hidelinks]{hyperref}
\usepackage{booktabs}

\newenvironment{subfig}[1]{\begin{subfigure}[t]{#1} }{\end{subfigure}}

\renewcommand{\vec}[1]{\boldsymbol{#1}}

\newcommand{\eg}{e.g., }
\newcommand{\ie}{i.e., }
\newcommand{\vs}{vs.}

\newcommand{\datasetname}{Active-Passive SimStereo}
\newcommand{\Datasetname}{ACTIVE-PASSIVE SIMSTEREO}

\begin{document}
\title{\datasetname{} - Benchmarking the Cross-Generalization Capabilities of Deep Learning-based Stereo Methods}

\author{
Laurent~Jospin$^{1*}$ \quad Allen~Antony$^1$ \quad Lian~Xu$^1$ \quad Hamid~Laga$^2$ \quad Farid~Boussaid$^1$ \\ \textbf{Mohammed~Bennamoun}$^1$ \\
$^1$University of Western Australia \quad $^2$Murdoch University\\
\texttt{\{laurent.jospin,lian.xu,farid.boussaid,mohammed.bennamoun\}@uwa.edu.au}\\
\texttt{H.Laga@murdoch.edu.au}
}

\maketitle

\begin{abstract}
In stereo vision, self-similar or bland regions can make it difficult to match patches between two images. Active stereo-based methods mitigate this problem by projecting a pseudo-random pattern on the scene so that each patch of an image pair can be identified without ambiguity. However, the projected pattern significantly alters the appearance of the image. If this pattern acts as a form of adversarial noise, it could negatively impact the performance of deep learning-based methods, which are now the de-facto standard for dense stereo vision. In this paper, we propose the \datasetname{} dataset and a corresponding benchmark to evaluate the performance gap between passive and active stereo images for stereo matching algorithms. Using the proposed benchmark and an additional ablation study, we show that the feature extraction and matching modules of a selection of twenty selected deep learning-based stereo matching methods generalize to active stereo without a problem. However, the disparity refinement modules of three of the twenty architectures (ACVNet, CascadeStereo, and StereoNet) are negatively affected by the active stereo patterns due to their reliance on the appearance of the input images.
\end{abstract}




\section{Introduction}

\begin{figure}[t]
    \centering
    
    \begin{subfig}{0.31\textwidth}
        \centering
        \includegraphics[width=\textwidth]{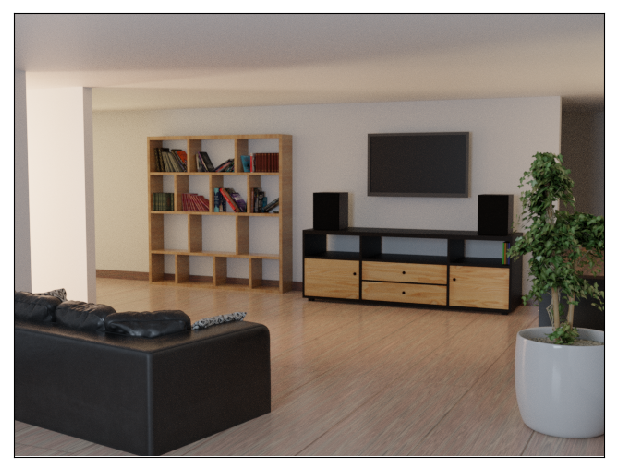}
        \caption{}
        \label{fig:datasetpreview:rgb}
    \end{subfig}
    \hfill
    \begin{subfig}{0.31\textwidth}
        \centering
        \includegraphics[width=\textwidth]{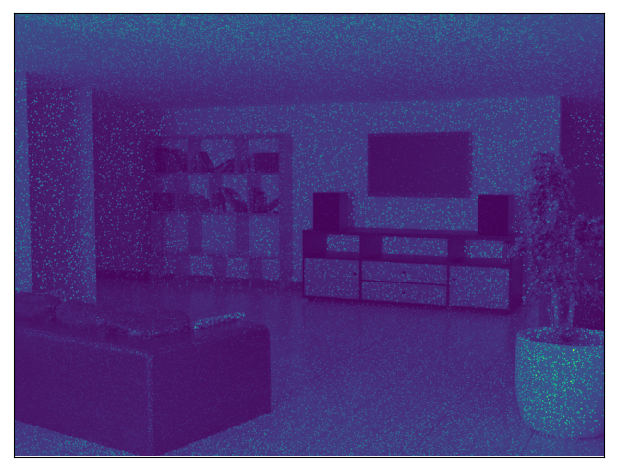}
        \caption{}
        \label{fig:datasetpreview:nir}
    \end{subfig}
    \hfill
    \begin{subfig}{0.31\textwidth}
        \centering
        \includegraphics[width=\textwidth]{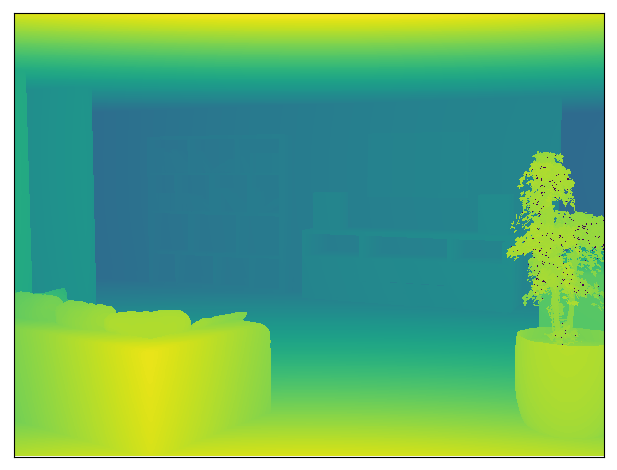}
        \caption{}
        \label{fig:datasetpreview:disp}
    \end{subfig}
    
    \caption{A sample from our dataset, with realistic (\subref{fig:datasetpreview:rgb}) passive and (\subref{fig:datasetpreview:nir}) active stereo images along with (\subref{fig:datasetpreview:disp}) their corresponding perfect ground truth disparities. The proposed dataset  allows the comparison of the relative performance of stereo vision methods when used for passive or active stereo matching.}
    \label{fig:datasetpreview}
\end{figure}

\begin{figure}[t]
\vspace{-5pt}
    \centering
    \includegraphics[width=0.7\textwidth]{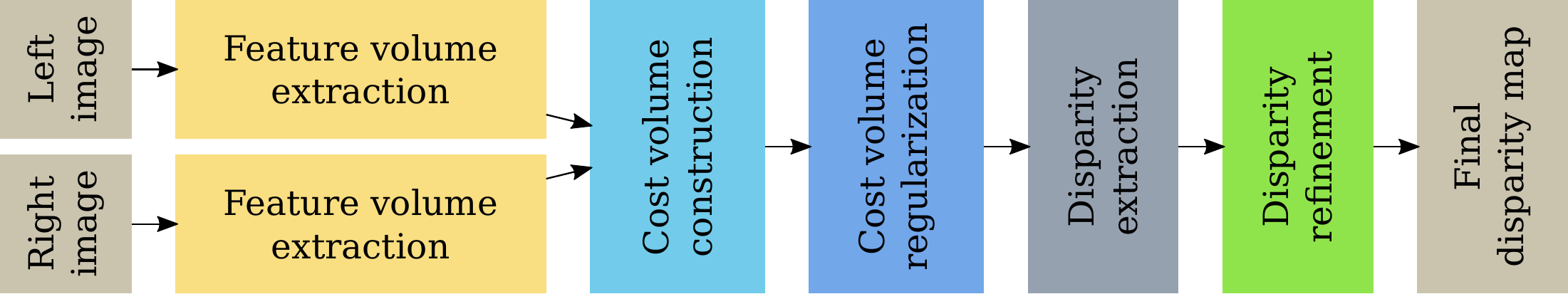}
    \caption{The typical stereo matching pipeline.}
    \label{fig:stereo_pipeline}
\vspace{-5pt}
\end{figure}

Stereo vision is used by many artificial or natural vision systems to acquire depth information from a pair of 2D projective views of the 3D world. In the context of computer vision, stereo matching operates in a multi-step pipeline (Fig.~\ref{fig:stereo_pipeline}) composed of: (\textbf{i}) a feature volume construction from the left and right views, (\textbf{ii}) a cost volume computation, which may be coupled with a regularization module, (\textbf{iii}) a disparity extraction from the cost volume, which is done using the argmin function, and (\textbf{iv}) a disparity refinement module, which may also use the cost volume and/or the image features as additional cues. The central step in this pipeline is the construction of the cost volume, which is  a function $C(x,y,d)$ that measures how unlikely a pixel of spatial coordinates $(x,y)$ is to be assigned a disparity value $d$.
Textureless and repetitive patterns in images can produce flat or periodic cost curves in the cost volume, leading to erroneous disparity maps in passive stereo systems, where only a pair of cameras is used. To address this issue, active stereo-based methods~\cite{Keselman_2017_CVPR_Workshops} project a pseudo-random light pattern on the scene to remove the textureless or self-similar areas in the stereo images  (Fig.~\ref{fig:datasetpreview}). Active stereo is now a critical component in many applications such as augmented reality \cite{holoportation2016} and robotics \cite{doi:10.1177/0278364911410755}. They are also part of consumer electronics devices such as smartphones \cite{10.1007/978-3-319-41501-7_50}.

Traditional stereo matching pipelines rely solely on closed-form formulations \cite{scharstein2002taxonomy}.
However, in recent years, learning-based methods have led to a series of breakthroughs in the field. Early learning-based methods focused on replacing one or more blocks in the traditional pipeline with a deep neural network. The latest methods, however, address the problem in an end-to-end fashion; see  \cite{laga2020survey} for a detailed survey. Due to the lack of public active stereo datasets and the fact that passive stereo was perceived as more challenging, most of these models have been trained for the passive stereo problem.
An important property of the closed-form formulae used in traditional stereo matching methods is that when non-self similar texture is added to the scene, their performance monotonically increases. This key feature is at the core of active stereo systems \cite{10.1117/12.7973334}. If one can determine that the latest deep learning methods can also leverage active pseudo random noise to improve their prediction, this would show that these methods are indeed learning to match similar regions of the images rather than fitting some bias into the data. Additionally, it provides some insight into the models' generalization ability, which is important for their safe deployment in their intended application (\eg autonomous driving).

Under ideal conditions, deep learning-based methods are expected to behave in a similar fashion to their non learning counterpart and exhibit improved performance when additional pseudo-random texture is added to the scene. Yet, many large-scale deep learning models see a degradation of their performances when used on datasets that are only slightly different from their original training datasets \cite{zhang2019domaininvariant}. They often require an adaptation procedure to generalize to new unseen domains \cite{Tonioni_2017_ICCV}. Furthermore, they can be severely affected by even little adversarial noise under certain circumstances \cite{szegedy2013intriguing}, as they are prone to overfitting on small biases present in their training data \cite{PAULLADA2021100336}. However, these flaws are not everywhere. For example, it has already been shown that once simulated images are close enough to real images, deep learning stereo systems generalize without issues \cite{Tosi2021CVPR}. Also, unlike adversarial noise, the pseudo random patterns used in active stereo have not been learned specifically to cause failure for deep learning models. This means that existing deep learning methods might generalize to the active stereo domain without any form of fine-tuning.

In this work, we investigate how different state-of-the-art deep learning-based stereo matching architectures are impacted when presented with active, instead of passive, stereo images. To make the evaluation of the generalization ability of stereo vision models easier, we propose \datasetname{}, a novel dedicated dataset composed of computer-generated images rendered using a physically-based rendering engine. The proposed dataset provides both active and passive frames for each given scene. This allows to evaluate and compare the performance of each algorithm on active  and passive stereo using exactly the same scenes.  The data set is publicly available at \url{https://dx.doi.org/10.21227/gf1e-t452}.

The remaining parts of the paper are organized as follows. Section~\ref{sec:related} reviews the related work. Section~\ref{sec:proposed_dataset} describes the proposed dataset. Section~\ref{sec:benchmarck} presents the proposed benchmark used for evaluation. Section~\ref{sec:results} presents and discusses the results of existing methods. Finally, Section~\ref{sec:conclusion} concludes the paper.

\section{Related Work}
\label{sec:related}

Many datasets and benchmarks have been proposed for passive stereo vision including the popular Middleburry dataset \cite{scharstein2002taxonomy, hirschmuller2007evaluation}, whose latest version uses a precise but expensive reconstruction pipeline to acquire the ground truth \cite{scharstein2014high}. The corresponding Middleburry Stereo Evaluation benchmark is widely used to evaluate stereo vision algorithms. 
Due to the challenges associated with  the 3D ground-truth acquisition, the aforementioned dataset only contains a small amount of labelled data, which is not sufficient to train large-scale deep architectures.
Subsequently, the Scene Flow datasets have been proposed \cite{mayer2016large}. They contain a large number of simulated image pairs with ground-truth optical flows and disparities generated from open source motion graphics short movies or randomized virtual 3D objects. However, the appearance of these simulated scenes is not realistic. Thus, most deep learning-based models for stereo vision need to be fine-tuned after being trained on the Scene Flow datasets. The UnrealStereo4K simulated dataset \cite{Tosi2021CVPR} was later proposed to provide higher resolution and more realistic images, taken from video games scenes. 

One of the most popular applications of stereo vision is autonomous driving, since vision based systems offer a cost-effective alternative or complement to LIDAR-based systems for depth measurement. Thus, many datasets and benchmarks have been specifically developed for this application. Examples include the KITTI Vision suite \cite{Menze2015CVPR, Geiger2012CVPR}, which is currently the most popular stereo vision benchmark for autonomous driving,  DrivingStereo \cite{Yang_2019_CVPR}, which is a large dataset commonly used for training rather than evaluation, and ApolloScape \cite{8753527}, which provides a benchmark suite for different challenges related to autonomous driving, including stereo vision. The ground truth of these datasets was obtained using a LIDAR-based system. Occasionally, a recognition system was also used to detect and categorize cars in images before aligning a CAD model onto the LIDAR depth map \cite{Menze2015CVPR, 8753527}. The inherent noise associated with these various processing steps implies that the ground truth cannot be trusted for very precise reconstructions. However, given that autonomous driving scenarios do accommodate a disparity error of one or two pixels, this is not a problem for the intended use of those datasets. 

For active stereo vision, there are far fewer public datasets, none of which has become popular for training or evaluating deep learning-based  stereo matching methods. The few end-to-end methods trained for active stereo use soft labels, \ie labels with associated uncertainty,  such as the depth generated by stereo cameras \cite{Zhang_2018_ECCV, 9360804}. Other methods did also use self-supervision, \eg by  using the information conserved when compressing a given image patch as supervisory signal \cite{Fanello_2017_CVPR}. 
Simulation techniques have also been proposed to generate semi-realistic images from CAD models \cite{8374552} based on screen space projection of texture. This approach has been used on multiple occasions \cite{Fanello_2017_CVPR, Zhang_2018_ECCV}, but none of the produced datasets has been made public. In this work, We use a similar approach  but with a physically-based rendering pipeline to improve the realism of the scenes, making our dataset more suitable for evaluation.

Datasets providing images for both active and passive stereo matching are even more scarce. To the best of our knowledge, only UnrealStereo4K \cite{Tosi2021CVPR} has monocular active frames for a subset of the images, but this part of the dataset has not been made publicly available. Furthermore, monocular active depth estimation is a slightly different problem from active stereo vision \cite{Riegler_2019_CVPR}, as the matching is performed between a pattern and an image, rather than between two images with a projected pattern. Thus, there exists no public dataset available for evaluating stereo models on active stereo images or evaluating the generalization capabilities of these models.

\section{The \Datasetname{} Dataset}
\label{sec:proposed_dataset}

\begin{figure}[tb]
    \centering
    \begin{subfig}{0.22\textwidth}
    \includegraphics[trim={0 50pt 0 0},clip,width=\textwidth]{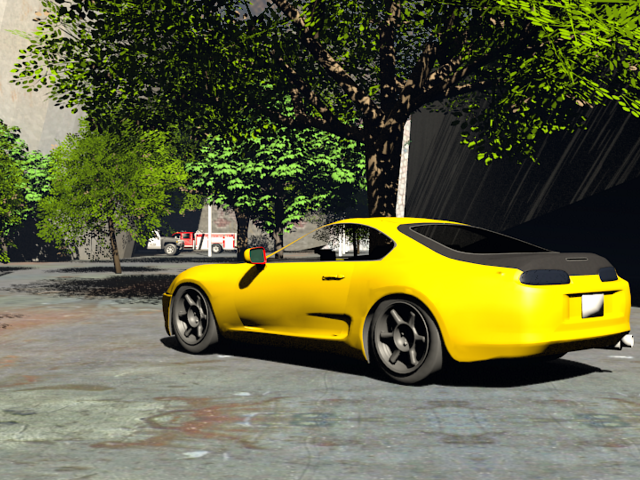}
    \caption{Sceneflow  \cite{mayer2016large}}
    \label{fig:realsim_comp:sceneflow}
    \end{subfig}
    \hfill
    \begin{subfig}{0.22\textwidth}
    \includegraphics[trim={0 50pt 0 0},clip,width=\textwidth]{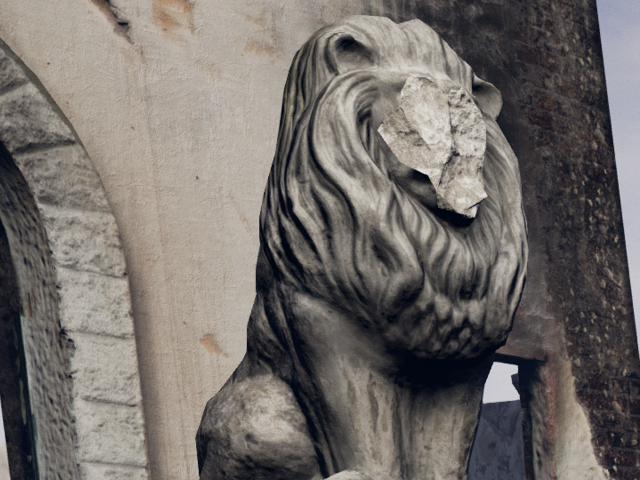}
    \caption{UnrealStereo4K \cite{Tosi2021CVPR}}
    \label{fig:realsim_comp:unreal4k}
    \end{subfig}
    \hfill
    \begin{subfig}{0.22\textwidth}
    \includegraphics[trim={0 50pt 0 0},clip,width=\textwidth]{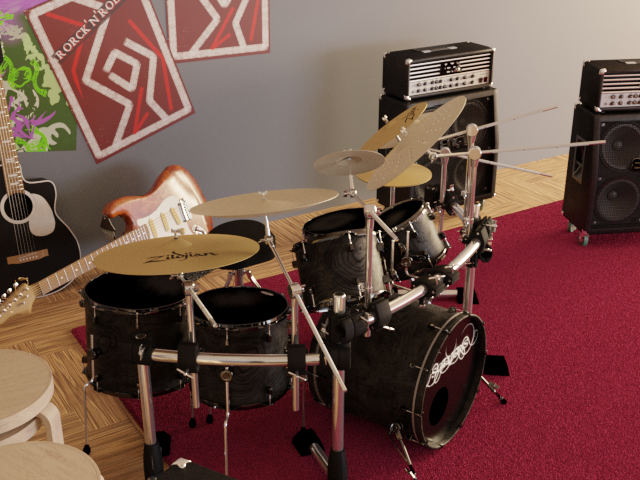}
    \caption{Ours}
    \label{fig:realsim_comp:music}
    \end{subfig}
    \hfill
    \begin{subfig}{0.22\textwidth}
    \includegraphics[trim={0 50pt 0 0},clip,width=\textwidth]{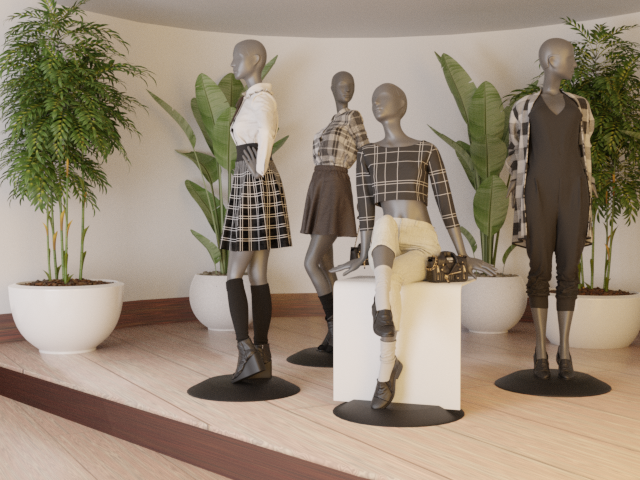}
    \caption{Ours}
    \label{fig:realsim_comp:mannequins}
    \end{subfig}
    
    \caption{Illustration of the benefits of a physically-based rendering pipeline.  In our images (\subref{fig:realsim_comp:music}), (\subref{fig:realsim_comp:mannequins}), indirect lighting creates soft and realist shadows and specularities, which is not the case with  existing simulated datasets (\subref{fig:realsim_comp:sceneflow}), (\subref{fig:realsim_comp:unreal4k}).}
    \label{fig:realsim_comp}
\end{figure}

Simulation offers both benefits and challenges for dataset creation. The size of real stereo datasets with high quality ground truth like Middlebury 2014 is limited because of the complex setups and amount of work needed \cite{scharstein2014high}. On the other hand, automated pipelines like the ones used for Kitti \cite{Menze2015CVPR}  are noisy. The Kitti benchmark is, therefore, limited to $BAD_N$ metrics (see Section~\ref{sec:benchmarck}) with large $N$ and is not suitable for subpixel accuracy comparisons. Simulation on the other hand makes labelling cheaper and noiseless.  It also allows to generate the exact same image twice, once for active stereo and once for passive stereo. This  greatly reduces biases when measuring the generalization ability of a given model, since the geometry does not change between frames. However, the simulation procedure may add a domain gap with real images. In computer vision, simulated data can be obtained using computer generated imagery (CGI). In recent years, CGI software has made tremendous advances. Using advanced CGI techniques, generating synthetic scenes that are indistinguishable from real images and match quite closely the acquisition process of real cameras became possible \cite{10.5406/jfilmvideo.69.1.0003}. This is done by using physically-based rendering (PBR), \ie a rendering engine simulating the physical behaviour of light \cite{pharr2016physically}. Figure~\ref{fig:realsim_comp} compares images from our dataset with images from other simulated datasets \cite{Tosi2021CVPR, mayer2016large}, which used real-time render instead of PBR. The most visible benefit of using PBR is that it produces more accurate indirect lighting, thereby creating softer and more accurate shadows. A better rendering of non-Lambertian materials \cite{pharr2016physically} is another benefit. This significantly reduces the domain gap between real and simulated data once a synthetic scene is created. In this paper, we purchased a series of high-quality realistic 3D assets to create realistic scenes for our dataset. Our dataset is not intended for training but for performance evaluation and fine-tuning, so we prioritize quality and diversity over quantity. To increase the number and the diversity of scenes, we also generated images that contain procedurally-generated shapes and images with abstract objects.

\subsection{Simulation Procedure}

\begin{figure}[t]
     \centering
    
    \begin{subfig}{0.45\textwidth}
    \includegraphics[width=\textwidth]{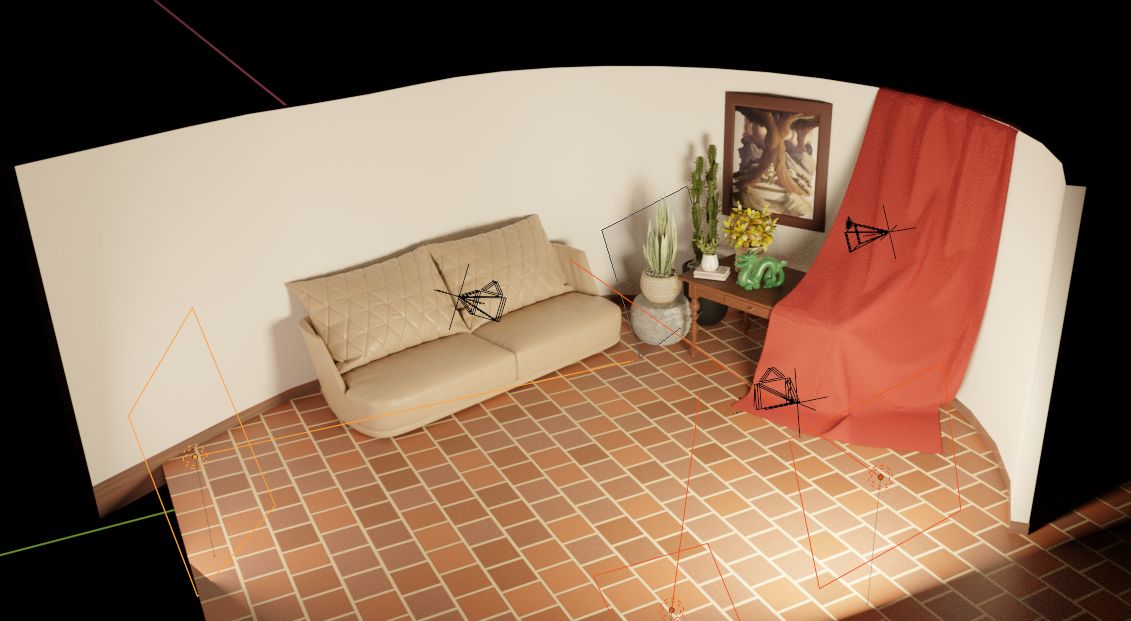}
    \caption{}
    \label{fig:simulationsetting:passive}
    \end{subfig}
    \hfill
    \begin{subfig}{0.45\textwidth}
    \includegraphics[width=\textwidth]{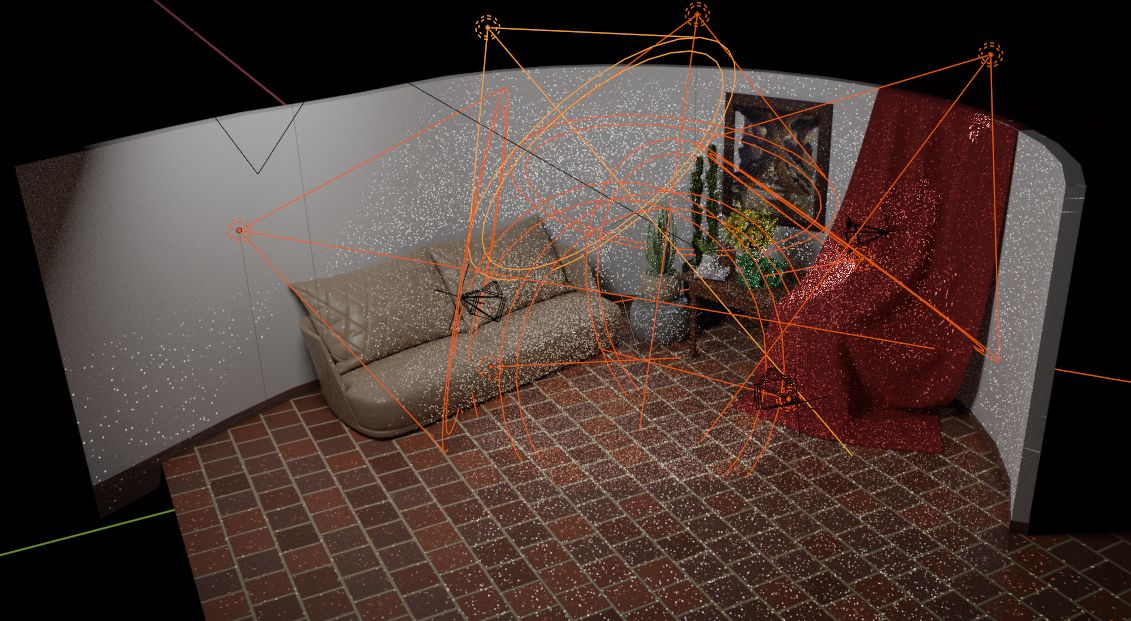}
    \caption{}
    \label{fig:simulationsetting:active}
    \end{subfig}
    \vspace{-5pt}
    
    \caption{The two different light setups to simulate (\subref{fig:simulationsetting:passive}) passive and (\subref{fig:simulationsetting:active}) active stereo acquisition in a given 3D scene.}
    \label{fig:simulationsetting}
    \vspace{-5pt}
\end{figure}

For each 3D scene, we designed two lighting setups. The first corresponds to a passive stereo acquisition scenario without any pseudo-random pattern while the second one corresponds to an active stereo acquisition scenario with a collection of lights projecting a pseudo-random pattern,  see Figure~\ref{fig:simulationsetting}).
We used the Cycles path tracing engine integrated into Blender \cite{BlenderSoftware} for rendering. We used standard shaders for the non-textured light sources. For the pseudo-random pattern light projectors, we used a custom programmed shader to generate a pattern resembling the one from the RealSense cameras. The light intensity $I$, is a function of the incoming direction $\vec{d}$:
\begin{equation}
\label{eq:patterngeneration}
    I(\vec{d}) = p \left((1-W_{s_1}(\vec{d} + \vec{t})^2)^{p_1} + c\right) \left(1-W_{s_2}(\vec{d} + \vec{t})^2\right)^{p_2}.
\end{equation}
\noindent Here, $W_{s_1}$ and $W_{s_2}$ are Whorley noise patterns ~\cite{worley1996cellular}; $s_1$ and $s_2$ are the scale factors of their respective Whorley noise pattern with $s_1 < s_2$; $p_1$ and $p_2$ are two light intensity correction factors with $p_1 \ll p_2$; $c$ is the minimal power of the $W_{s_2}$ pattern; $\vec{t}$ is a random translation of the texture space to generate different patterns for different lights; and $p$ is the power gain of the lamp, expressed in Watts.

%

The ground truth is then extracted from the depth pass $z$ generated by the rendering software. $z$ measures the distance between the visible point in the 3D scene and the optical center of the camera. The ground truth disparity $\check{d}$ can be computed as:
\begin{equation}
    \check{d} = \frac{B f}{z}, 
\end{equation}
\noindent where $B$ is the stereo camera baseline and $f$ is the camera focal length in pixel. We used a Baseline of $0.16$m (where m here refers to the unit used internally by Blender, not the actual meters in the real world), and cameras with a focal length of $48.61$mm, which amounts to $888.89$ pixels at standard resolution for a $35$mm film equivalent sensor.


    
%

\begin{figure*}[t]
    \centering
    \includegraphics[width=\textwidth]{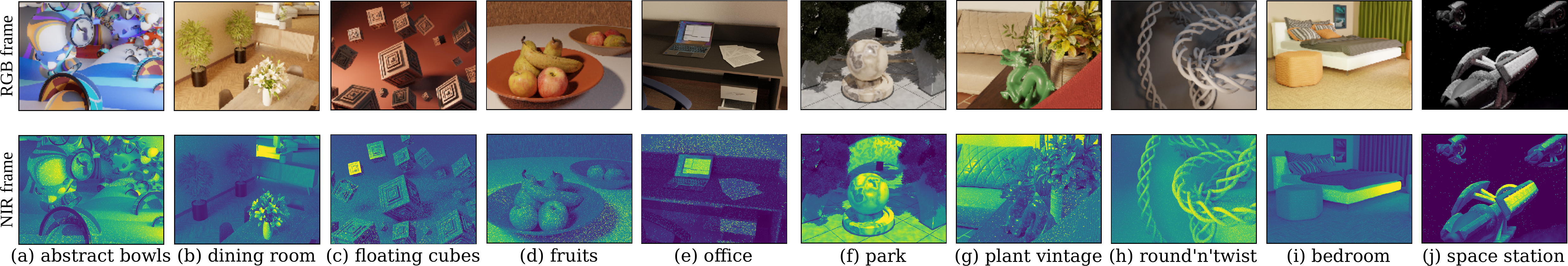}
    \caption{Sample images from the proposed test set.}
    \label{fig:test_set}
\end{figure*}

\subsection{The Dataset}

The proposed dataset contains $515$ image pairs split into a training set ($80\%$ of the images) and a test set ($20\%$ of the images) used for benchmarking. The test set contains $103$ image pairs (Fig.~\ref{fig:test_set}), comprising different shapes (\eg large flat surfaces such as floors or small areas with depth discontinuities such as plant leaves), depth ranges, and styles (\ie realistic scenes or abstract compositions). The remaining $412$ image pairs are contributed as a training set. We use the standard resolution ($640\times480$ pixels) for the benchmark as it matches or approaches the resolution of most stereo cameras. It also  guarantees that the images can be processed by most methods even on memory-constrained hardware. Despite the small number of images, the test set is shown to be large enough to evaluate deep learning methods. Detailed experiments to demonstrate this are provided in the Supplementary Material. Our dataset is also large enough for fine-tuning (see Section~\ref{sec:finetune}). 

In addition to the benchmark, we provide a simulation Blender file with the specific shader, as well as the python code to post-process the images.

\begin{table}[htb]

    \caption{Comparison of the resolution and disparity ranges of different datasets.}
    \label{table:datasets_infos}
    \vspace{5pt}
    \centering
    \resizebox{\columnwidth}{!}{
    \begin{tabular}{lrrrrrr}
    \toprule
    & & & & \multicolumn{3}{l}{Disparity {[}px{]}} \\
    \cmidrule(r){5-7}
    Dataset & Generation method & \# images pairs & Resolution {[}px{]} & Min & Mean & Max \\ \midrule
    Ours (train) & Ray tracing rendering & 412 & $640 \times 480$ & 0.00 & 21.77 & 212.67 \\ 
    Ours (test) & Ray tracing rendering & 103 & $640 \times  480$ & 0.00 & 25.12 & 129.67 \\ 
    Middlebury 2014 \cite{scharstein2014high} & Large baseline active stereo & 33 & $2850 \times  1900$\textsuperscript{1} & 28.94 & 148.36 & 695.61 \\ 
    Kitti 2012 \cite{Geiger2012CVPR} & Real images + Laser scanner & 389 & $1240 \times  380$\textsuperscript{1} & 4.11 & 38.32 & 227.99 \\ 
    Kitti 2015 \cite{Menze2015CVPR} & Real images + Laser scanner + 3D cad object alignment & 400 & $1240 \times  380$\textsuperscript{1} & 4.46 & 33.62 & 229.96 \\ 
    Sceneflow  \cite{mayer2016large} & Screen space rasterization & 39049 & $960 \times 540$ & 1.12 & 39.87 & 940.75 \\ 
    UnrealStereo4k \cite{Tosi2021CVPR} & Screen space rasterization & 7200 & $3840 \times  2160$ & 0.01 & 173.20 & 1515.60 \\ \bottomrule
    \end{tabular}}
    
    \caption*{\small \textsuperscript{1} The images in these datasets have variable sizes. Thus, the values given here are an approximation.}
    
\end{table}

%
%

\section{The Benchmark}
\label{sec:benchmarck}

We use five different scores to compare the generalisation abilities of different methods. The most important metric in stereo vision is the BAD-N, which measures the proportion of pixels above a given error threshold $N$.  It is computed over a test set $T$ as:
\begin{equation}
    BAD_N = \frac{1}{\left\|T\right\|} \sum_{I \in T} \frac{\sum_{i = 0}^{h_I} \sum_{j = 0}^{w_I} \mathbf{1}_{\left| \Delta d_{i,j} \right| > N } }{h w},
\end{equation}
\noindent where $h$ and $w$, respectively, are the height and the width of the image $I$, $\mathbf{1}_{A}$ is the indicator function of $A$ and $\Delta d$ is the disparity error for the image $I$. We report results for $N \in \{0.5, 1, 2, 4\}$. Lower BAD$_N$ scores indicate a better accuracy in reconstructing the disparity. In the rest of the paper, we will indicate this by appending a down arrow ($\downarrow$) after the name of these metrics.

We also use the Mean Absolute Error (MAE) and the Root Mean Square Error (RMSE). The former is a good estimate of the expected amplitude of the error of a given method, and the latter, especially when compared to the MAE, is a good indicator of the presence of outlier points with large errors. The RMSE and MAE scores over $T$ are computed as:
\begin{equation}
    RMSE_T = \frac{1}{\left\| T \right\|} \sum_{I \in T} \sqrt{\frac{\sum_{i = 0}^{h} \sum_{j = 0}^{w} \Delta d_{i,j}^2}{{h  w}} }, ~~~  MAE_T = \frac{1}{\left\| T \right\|} \sum_{I \in T} \frac{\sum_{i = 0}^{h} \sum_{j = 0}^{w} \left| \Delta d_{i,j} \right|}{{h w}} ~.
\end{equation}

\noindent These scores are the average of the corresponding metric over the test set. Lower MAE and RMSE indicate a better accuracy in reconstructing the disparity. In the rest of the paper, we will indicate this by appending a down arrow ($\downarrow$) after the name of these metrics.

A consequence of using the MAE and RMSE is that if a method exhibits low performances on a specific image, the overall score of the method will be greatly influenced by this single image. For an absolute performance evaluation benchmark, this is not a problem. However,  this is an issue when evaluating the relative performance variation resulting from a domain change (\eg from passive stereo to active stereo). To mitigate this use, we measure the mean relative score variation across all images.
Given a metric $M$, the relative score variation $R_M$ is computed as:
\begin{equation}
\label{eq:relscorevar}
    R_M = \frac{1}{\left\| T \right\|} \sum_{I \in T} \frac{M_{I_P} - M_{I_A}}{M_{I_P}} ~,
\end{equation}
\noindent where $M_{I_P}$ and $M_{I_A}$ are the metric scores evaluated on the passive stereo results and active stereo results of an image $I$, respectively. In this paper, we focus on $R_{MAE}$ and $R_{BAD_2}$.

We also report the proportion $P_{M}$ of the testing images in which the active stereo results outperform their passive stereo counterparts in terms of the metric $M$. It is formulated as:
\begin{equation}
\label{eq:propimpimgs}
     P_{M} = \frac{1}{\left\| T \right\|} \sum_{I \in T} \mathbf{1}_{M_{I_A} < M_{I_P}}.
\end{equation}
\noindent In this paper, we focus on $P_{MAE}$ and $P_{BAD_2}$. We report in the supplementary material  other variants of those metrics.

Higher P and R scores indicate a better generalization of the method from passive stereo to active stereo. In the rest of the paper, we will indicate this by appending an up arrow ($\uparrow$) after the name of these metrics.

Finally, discontinuities in the depth image are an important parameter in evaluating stereo reconstruction methods. Discussing the detailed performances of existing methods would be beyond the scope of this paper, which focuses on the generalization capabilities from passive to active stereo and vice versa. Nonetheless, we also included, in the supplementary material, versions of all the metrics presented above, restricted either on the edge regions of the image or on its flat regions.

\section{Results on Existing Methods}
\label{sec:results}

\begin{table}[]
    \centering
    \caption{Evaluated methods with training sets and results on public benchmarks}
    \label{tab:methods_overview}
    \resizebox{\columnwidth}{!}{
    \begin{tabular}{llrrrrrrr}
        \toprule
        \multicolumn{2}{c}{ } & \multicolumn{2}{c}{Training for our evaluation} & & \multicolumn{1}{c}{KITTI 2015 \cite{Menze2015CVPR}} & & \multicolumn{2}{c}{Middleburry \cite{scharstein2014high}} \\
        \cmidrule{3-4}\cmidrule{6-6}\cmidrule{8-9}
        Method & Stereo type & Train set & Fine-tuned set & & D1-all $\downarrow$ & & BAD$_2$ $\downarrow$ & MAE $\downarrow$\\ \midrule
        AANet \cite{xu2020aanet} & Passive & SceneFlow & KITTI 2015 & & 2.55\% & & 25.20\% & 8.88px \\
        ACVNet \cite{xu2022attention} & Passive & SceneFlow & KITTI 2015 & & 1.65\% & & 13.60\% & 8.24px \\
        AnyNet \cite{wang2018anytime} & Passive & SceneFlow & KITTI 2015 & & 6.20\% & & - & - \\
        CascadeStereo \cite{gu2019cas} & Passive & SceneFlow & KITTI 2015 & & 2.00\% & & 18.80\% & 4.50px \\
        CREStereo \cite{li2022practical} & Passive & CREStereo & ETH3D & & 1.69\% & & 3.71\% & 1.15px \\
        Deep-Pruner (best) \cite{9009458} & Passive & SceneFlow & KITTI 2015 & & 2.15\% & & 30.10\% & 4.80px\\
        Deep-Pruner (fast) \cite{9009458} & Passive & SceneFlow & KITTI 2015 & & 2.59\% & & - & -\\ 
        GA-Net \cite{Zhang2019GANet} & Passive & SceneFlow & KITTI 2015 & & 2.59\% & & 18.90\% & 12.20px \\ 
        GwcNet \cite{guo2019group} & Passive & SceneFlow & KITTI 2015 & & 2.11\% & & - & - \\ 
        HighResStereo \cite{yang2019hsm} & Passive & SceneFlow & KITTI 2015 & & 2.14\% & & 10.20\% & 2.07px \\ 
        Lac-GwcNet \cite{Liu_Yu_Long_2022} & Passive & SceneFlow & KITTI 2015 & & 1.77\% & & - & - \\ 
        MobileStereoNet3D \cite{shamsafar2022mobilestereonet} & Passive & SceneFlow & KITTI 2015 & & 2.10\% & & - & - \\
        MobileStereoNet2D \cite{shamsafar2022mobilestereonet} & Passive & SceneFlow & KITTI 2015 & & - & & - & - \\
        PSM-Net \cite{chang2018pyramid} & Passive & SceneFlow & KITTI 2015 & & 2.32\% & & 42.10\% & 6.68px \\
        RAFT-Stereo \cite{lipson2021raft} & Passive & SceneFlow & - & & - & & 4.74\% & 1.27px \\
        RealTimeStereo \cite{Chang_2020_ACCV} & Passive & SceneFlow & KITTI 2015 & & 7.54\% & & - & - \\
        SMD-Nets \cite{Tosi2021CVPR} & Passive & UnrealStereo4K & KITTI 2015 & & 2.08\% & & - & - \\
        SRH-Net \cite{Hongzhi2021SRH} & Passive & SceneFlow & KITTI 2015 & & - & & - & - \\
        StereoNet \cite{khamis2018stereonet} & Passive & SceneFlow & - & & - & & - & - \\
        ActiveStereoNet \cite{Zhang_2018_ECCV} & Active & Self-supervised & - & & - & & - & - \\ \bottomrule
    \end{tabular}}
\end{table}

We used the proposed benchmark to evaluate state-of-the-art, end-to-end deep neural networks for stereo-matching. We considered $20$ methods for which the source code and the pre-trained models are available. In the supplementary material, we also evaluate a selection of traditional non-learning methods. Those methods are listed in Table~\ref{tab:methods_overview}, along with the datasets used to train the models. We used mainly the models weights trained on SceneFlow \cite{mayer2016large} and fine-tuned on KITTI 2015 \cite{Menze2015CVPR} as this approach is the standard for deep stereo models \cite{laga2020survey}. For each method, we list the reported D1 score in the KITTI 2015 benchmark \cite{Menze2015CVPR}. This is the proportion of pixels with an error greater than 3px or 5\% of the disparity. We also report, if available, the BAD$_2$ and MAE scores of the method in the Middlebury benchmark \cite{scharstein2014high}.
    
\noindent \emph{As we are more interested in the generalization abilities of the different methods than their peak performances, we did not fine-tune any of these methods to our dataset.} This is to ensure the measured generalization performances are not biased by fine-tuning for one specific modality in our dataset. On the other hand, fine-tuning for both modalities   would not give the same insights on the generalization abilities of each method. However, this means that the reported performances should not be considered as the best achievable performances of the studied methods. In a second step, we fine-tune the methods which failed to generalize on our dataset to check if it can be used to fine-tune passive stereo methods on active stereo (Section~\ref{sec:finetune}). 


%
%
%

In this paper, we focus on the main aggregate results as well as on our general observations when analysing the error maps. The detailed scores per image are provided in the supplementary material.

\begin{table}[ht]
    \begin{center}
\caption{Evaluation results of state-of-the-art methods on passive and then active stereo images.}
    \label{tab:passive_dataset_results1}
    \vspace{5pt}
    \resizebox{\columnwidth}{!}{
    \begin{tabular}{lrrrrrrcrrrrrr}
        \toprule
        & \multicolumn{6}{c}{Passive stereo images} & & \multicolumn{6}{c}{Active stereo images}\\
        \cmidrule{2-7}\cmidrule{9-14}
        Method          & RMSE$_T$ $\downarrow$    & MAE$_T$ $\downarrow$    & BAD$_{0.5}$ $\downarrow$ & BAD$_1$ $\downarrow$  & BAD$_2$ $\downarrow$  & BAD$_4$ $\downarrow$  & & RMSE$_T$ $\downarrow$  & MAE$_T$ $\downarrow$    & BAD$_{0.5}$ $\downarrow$ & BAD$_1$ $\downarrow$ & BAD$_2$ $\downarrow$ & BAD$_4$ $\downarrow$ \\ \midrule
        AANet \cite{xu2020aanet} & 8.35px & 4.07px & 66\% & 45\% & 30\% & 20\% & & 6.73px & 2.02px & 43\% & 20\% & 12\% & 8\% \\
        ACVNet \cite{xu2022attention} & 9.22px & 4.00px & 36\% & 25\% & 17\% & 12\% & & 3.49px & 1.31px & 23\% & 13\% & 8\% & 5\% \\
        AnyNet \cite{wang2018anytime} & 8.80px  & 5.31px & 84\%   & 69\% & 46\% & 29\% & & 6.13px & 3.43px & 80\%   & 61\% & 32\% & 18\% \\
        Cascade-Stereo \cite{gu2019cas}  & 10.12px & 4.95px & 66\%   & 39\% & 23\% & 16\% & & 4.48px & 2.07px & 64\%   & 33\% & 14\% & 8\% \\
        CREStereo \cite{li2022practical} & 1.75px &  0.71px & 21\% & 14\% & 8\% & 4\% & & 1.44px  & 0.32px & 7\% & 4\% & 2\% & 1\% \\
        Deep-Pruner (best) \cite{9009458}     & 7.07px  & 3.80px & 50\%   & 34\% & 24\% & 17\% & & 3.16px & 1.12px & 23\%   & 12\% & 7\% & 5\%  \\
        Deep-Pruner (fast) \cite{9009458}     & 7.97px  & 4.73px & 67\%   & 47\% & 32\% & 22\% & & 4.27px & 1.91px & 41\%   & 23\% & 13\% & 9\%  \\
        GA-Net \cite{Zhang2019GANet} & 7.89px & 4.01px & 59\%   & 39\% & 27\% & 19\% & & 5.05px & 1.56px & 27\%   & 14\% & 9\% & 7\%  \\
        GwcNet \cite{guo2019group} & 10.38px & 5.17px & 74\% & 50\% & 33\% & 22\% & & 4.41px & 1.80px & 53\% & 24\% & 11\% & 8\% \\
        High-Res-Stereo \cite{yang2019hsm} & 4.87px  & 2.94px & 54\%   & 37\% & 24\% & 16\% & & 3.11px & 1.30px & 38\%   & 20\% & 10\% & 5\%  \\
        Lac-GwcNet \cite{Liu_Yu_Long_2022} & 7.71px & 3.95px & 58\% & 39\% & 26\% & 18\% & & 3.34px & 1.25px & 31\% & 14\% & 8\% & 5\% \\
        MobileStereoNet3D \cite{shamsafar2022mobilestereonet} & 9.81px & 5.06px & 81\% & 61\% & 38\% & 23\% & & 4.44px & 2.11px & 70\% & 38\% & 14\% & 8\% \\
        MobileStereoNet2D \cite{shamsafar2022mobilestereonet} & 7.68px & 4.49px & 79\% & 59\% & 36\% & 23\% & & 4.37px & 1.88px & 59\% & 28\% & 13\% & 8\% \\
        PSM-Net \cite{chang2018pyramid}  & 6.57px  & 4.23px & 93\%   & 82\% & 45\% & 21\% & & 3.94px & 2.32px & 93\%   & 80\% & 27\% & 7\%  \\
        RAFT-Stereo \cite{lipson2021raft} & 2.20px & 0.92px & 27\% & 17\% & 10\% & 5\% & & 1.68px & 0.47px & 13\% & 7\% & 3\% & 2\% \\
        RealTimeStereo \cite{Chang_2020_ACCV} & 7.69px & 4.71px & 80\% & 64\% & 44\% & 28\% & & 5.30px & 2.87px & 72\% & 50\% & 28\% & 15\% \\
        SMD-Nets \cite{Tosi2021CVPR} & 12.32px & 6.48px & 77\% & 58\% & 40\% & 27\% & & 5.26px & 2.24px & 63\% & 34\% & 15\% & 9\% \\
        SRH-Net \cite{Hongzhi2021SRH} & 7.43px & 4.24px & 76\% & 49\% & 29\% & 19\% & & 3.95px & 1.59px & 57\% & 21\% & 9\% & 5\% \\
        StereoNet \cite{khamis2018stereonet} & 10.98px & 4.11px & 55\%   & 37\% & 26\% & 18\% & & 7.74px & 1.79px & 44\%   & 23\% & 12\% & 7\%  \\
        ActiveStereoNet \cite{Zhang_2018_ECCV} & 21.57px & 9.39px & 60\%   & 46\% & 35\% & 28\% & & 6.92px & 2.32px & 36\%   & 22\% & 14\% & 9\% \\ \bottomrule
    \end{tabular}}
    \end{center}
\end{table}

\begin{table}[ht]
    \begin{center}
    \caption{Relative scores (equations~\ref{eq:relscorevar} and \ref{eq:propimpimgs}) for the state-of-the-art methods.}
    \label{tab:active_relative_scores}
    \vspace{5pt}
    \scalebox{0.8}{
    \begin{tabular}{lrrrr}
        \toprule
        Method          & $P_{MAE}$ $\uparrow$ & $P_{BAD_2}$ $\uparrow$ & $R_{MAE}$ $\uparrow$ & $R_{BAD_2}$ $\uparrow$ \\ \midrule
        AANet \cite{xu2020aanet} & 87\% & 99\% & 35\% & 55\% \\
        ACVNet \cite{xu2022attention} & 71\% & 74\% & 30\% & 37\% \\
        AnyNet \cite{wang2018anytime} & 93\% & 98\% & 29\% & 29\% \\
        Cascade-Stereo \cite{gu2019cas} & 65\% & 71\% & 25\% & 28\% \\
        CREStereo \cite{li2022practical} & 80\% & 62\% & 41\% & 34\% \\
        Deep-Pruner (best) \cite{9009458} & 94\% & 100\% & 52\% & 62\% \\
        Deep-Pruner (fast) \cite{9009458} & 95\% & 98\% & 47\% & 56\% \\
        GA-Net \cite{Zhang2019GANet} & 87\% & 96\% & 43\% & 59\% \\
        GwcNet \cite{guo2019group} & 99\% & 99\% & 55\% & 63\% \\
        High-Res-Stereo \cite{yang2019hsm} & 84\% & 85\% & 36\% & 49\% \\
        Lac-GwcNet \cite{Liu_Yu_Long_2022} & 96\% & 96\% & 55\% & 63\% \\
        MobileStereoNet3D \cite{shamsafar2022mobilestereonet} & 96\% & 99\% & 45\% & 59\% \\
        MobileStereoNet2D \cite{shamsafar2022mobilestereonet} & 96\% & 99\% & 52\% & 65\% \\ 
        PSM-Net \cite{chang2018pyramid} & 90\% & 93\% & 30\% & 38\% \\
        RAFT-Stereo \cite{lipson2021raft} & 81\% & 73\% & 37\% & 39\% \\
        RealTimeStereo \cite{Chang_2020_ACCV} & 92\% & 97\% & 34\% & 35\% \\
        
        SMD-Nets \cite{Tosi2021CVPR} & 95\% & 97\% & 54\% & 60\% \\
        SRH-Net \cite{Hongzhi2021SRH} & 97\% & 98\% & 48\% & 64\% \\
        
        StereoNet \cite{khamis2018stereonet} & 84\% & 85\% & 38\% & 45\% \\
        ActiveStereoNet \cite{Zhang_2018_ECCV} & 99\% & 99\% & 68\% & 58\% \\
        \bottomrule
    \end{tabular}}
    \end{center}
\end{table}

Table~\ref{tab:passive_dataset_results1} reports results on passive and active stereo images. We observed that, when evaluated on active stereo images, all considered methods show an improvement in their performance for all considered metrics. \emph{Please keep in mind that not all methods have been trained on the same dataset or even domain; our benchmark aims to evaluate the relative performances of the different methods rather than their absolute performances.}
ActiveStereoNet \cite{Zhang_2018_ECCV} stands out as the worst performing method for passive stereo but its performance drastically improves when presented with active stereo images. This is due to the fact that it is the only one trained on active stereo images. A model trained for active stereo is not expected to generalize well on passive stereo without adaptation since the domain shift makes matching much harder. ActiveStereoNet is far from being the best performing model on active stereo images despite being the only one trained on that domain. The neural network architecture is probably at play here. ActiveStereoNet uses the same architecture as StereoNet \cite{khamis2018stereonet}, albeit a different training method. StereoNet is not the best performing method on passive stereo and is outperformed by ActiveStereoNet on active stereo images, which demonstrates the benefits of the training strategy proposed by the authors of ActiveStereoNet.

The relative score improvements, reported in Table~\ref{tab:active_relative_scores}, also show that, overall, existing methods have a good capability to generalize to active stereo. ACVNet \cite{xu2022attention}, Cascade Stereo  \cite{gu2019cas} as well as CREStereo \cite{li2022practical} and RAFT-Stereo \cite{lipson2021raft} are the only methods that seem to have issues generalizing, as their P$_{MAE}$ and P$_{BAD_2}$ scores are way lower than the other methods. For CREStereo \cite{li2022practical} and RAFT-Stereo \cite{lipson2021raft}, it turns out this is more due to the fact that these methods perform so well on passive stereo, see Table~\ref{tab:passive_dataset_results1}, that they have less room for improvement when moving to active stereo. For ACVNet \cite{xu2022attention} and Cascade Stereo \cite{gu2019cas}, this is because moving to the active stereo domain causes artifacts in the reconstructed disparities, see Section~\ref{sec:ablstudy} for more details.

These results are  encouraging and show that, when trained on passive stereo, current state-of-the-art deep learning methods are able to generalize to active stereo. Generalizing to passive stereo for a method trained on active stereo, seems to be much harder. However, the results are not very conclusive, since only one method is covered, and it has been trained in a self-supervised fashion.

The details of these results, as well as the performance of each method on each image, are provided in the results Excel sheet included in the supplementary material.

\section{Ablation study}
\label{sec:ablstudy}

\begin{figure*}[t]
    \centering
    
    \begin{subfigure}{0.31\textwidth}
        \centering
        \includegraphics[width=\textwidth]{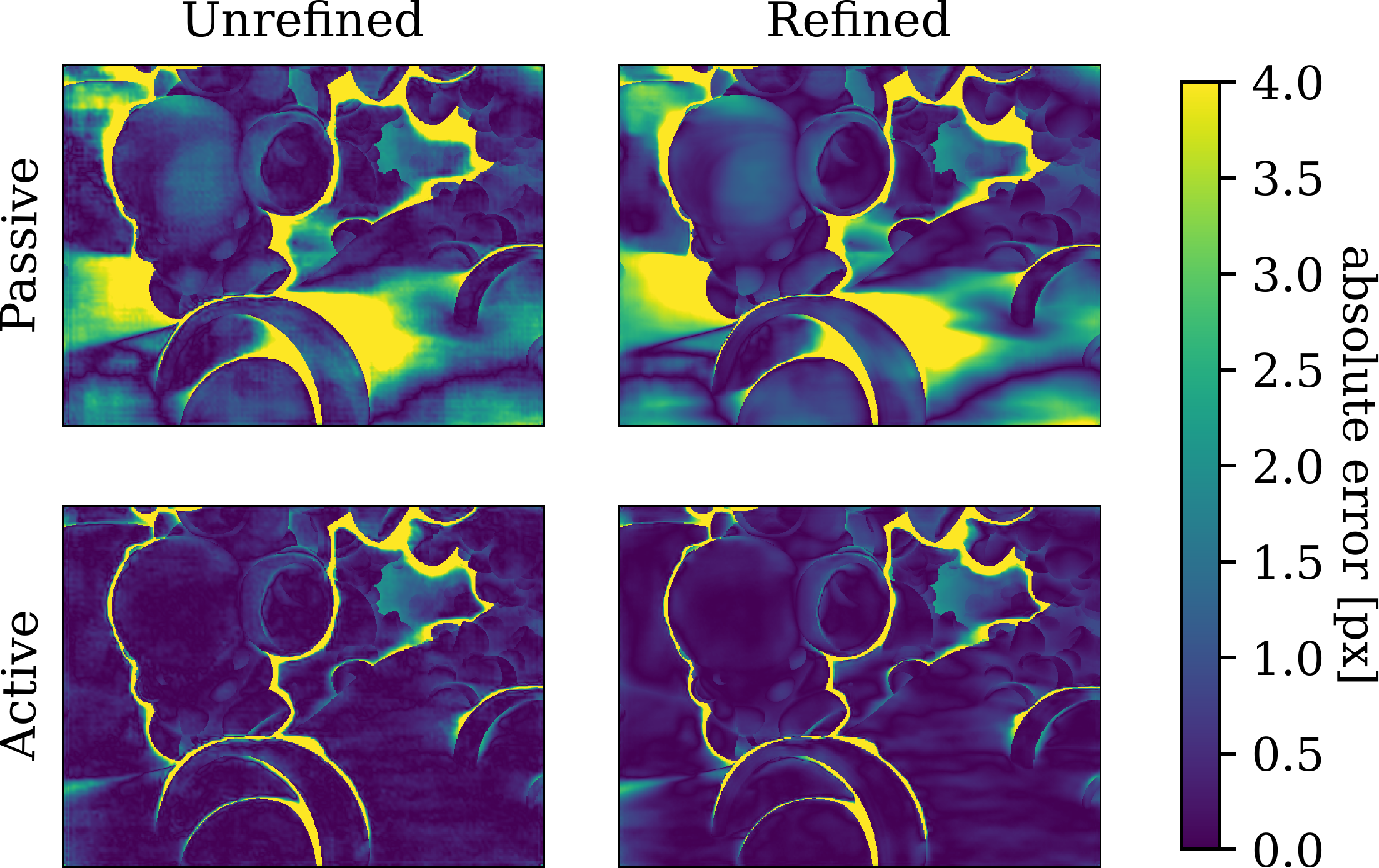}
        \caption{Abstract bowls of Fig.~\ref{fig:test_set}a with Deep-Pruner (best) \cite{9009458}.}
        \label{fig:errormapsclean:dp}
    \end{subfigure}
    \hfill
    \begin{subfigure}{0.31\textwidth}
        \centering
        \includegraphics[width=\textwidth]{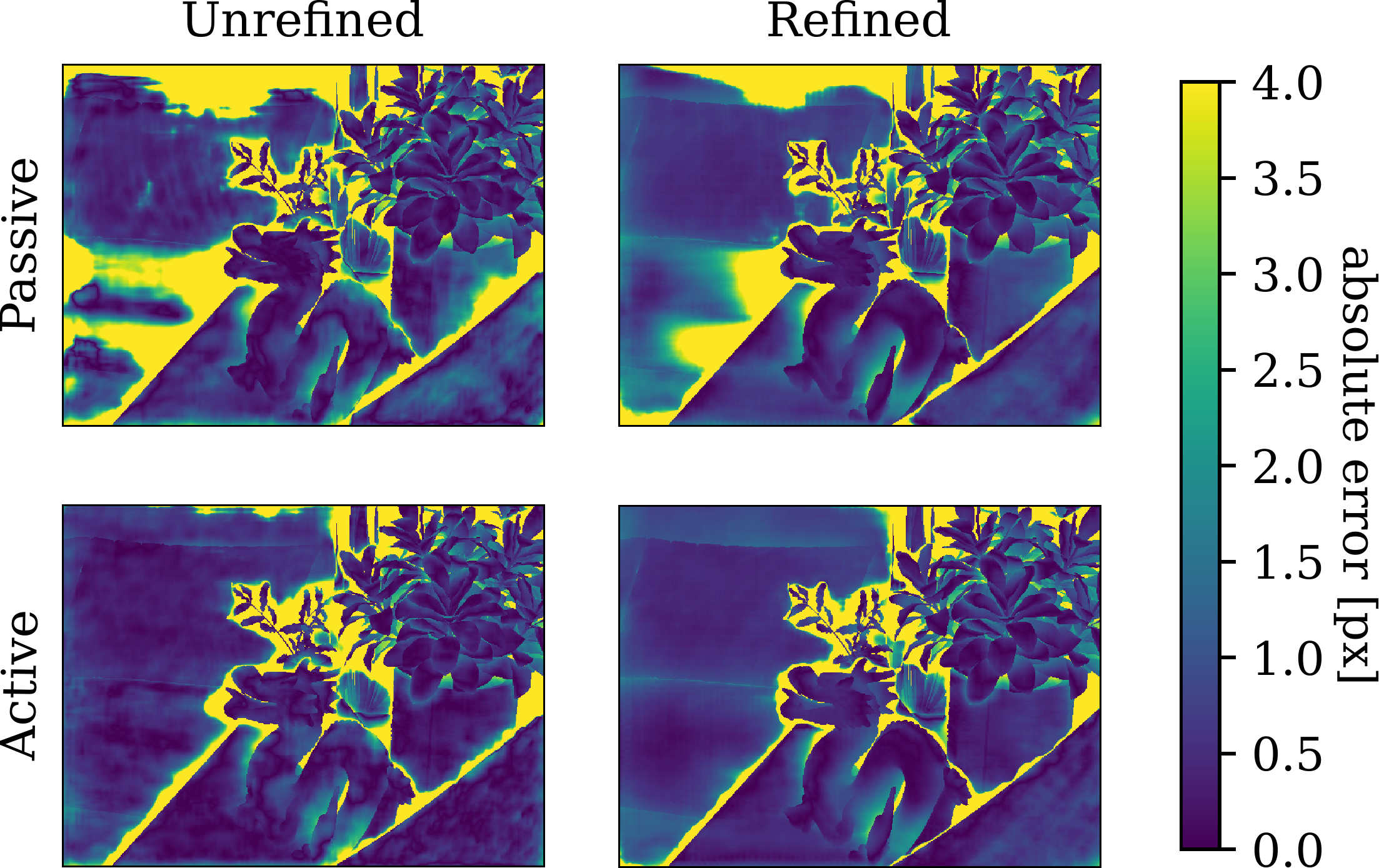}
        \caption{Plant vintage of Fig.~\ref{fig:test_set}g with GwcNet \cite{guo2019group}.}
        \label{fig:errormapsclean:gwc}
    \end{subfigure}
    \hfill
    \begin{subfigure}{0.31\textwidth}
        \centering
        \includegraphics[width=\textwidth]{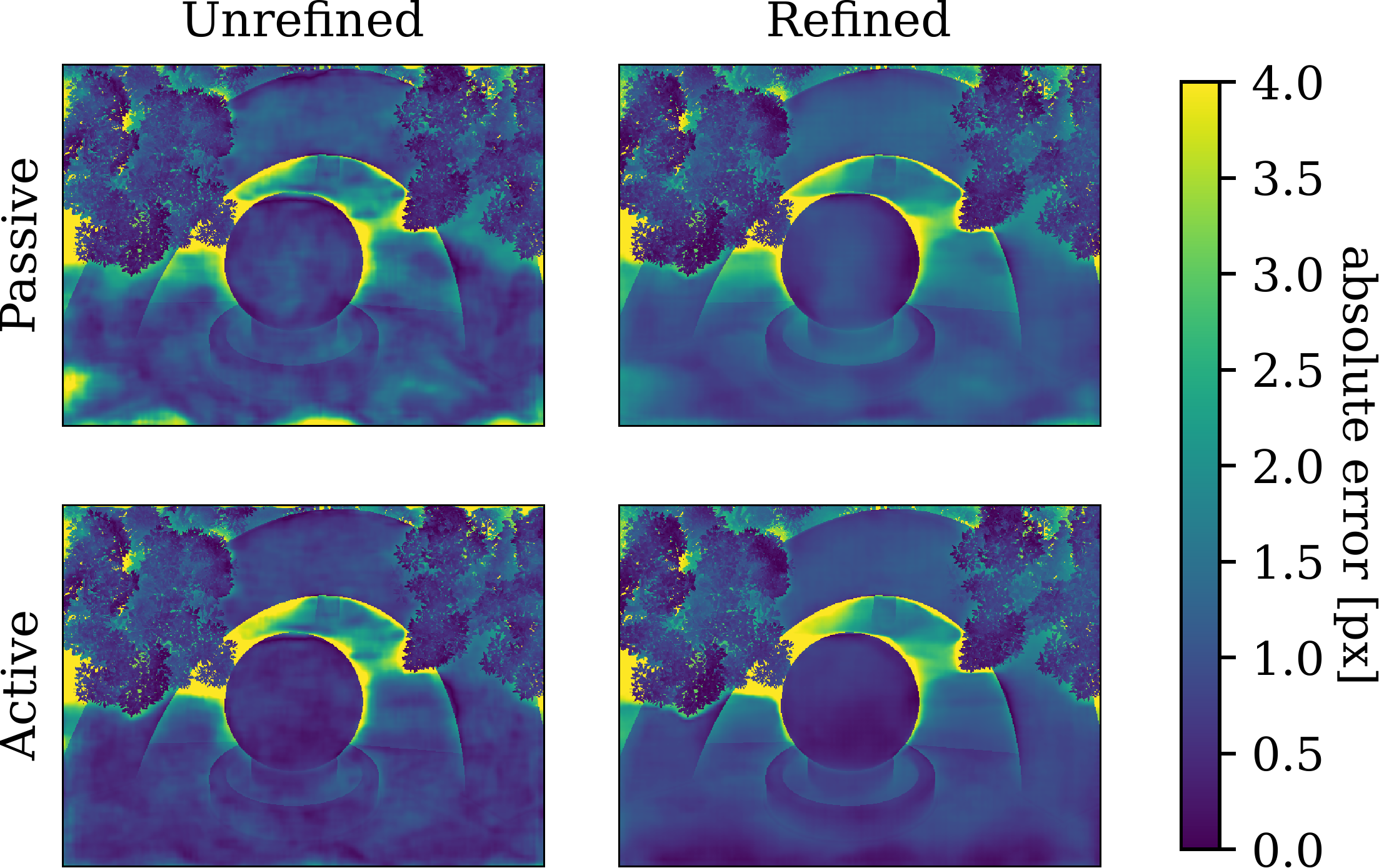}
        \caption{Park of Fig.~\ref{fig:test_set}f with MobileStereoNet3D \cite{shamsafar2022mobilestereonet}.}
        \label{fig:errormapsclean:mobile}
    \end{subfigure}
    
    \caption{Detailed error maps comparison between the unrefined and refined disparity maps for methods without apparent problems in active stereo compared to passive stereo.}
    \label{fig:errormapsclean}
\end{figure*}

\begin{figure*}[t]
    \centering
    
    \begin{subfigure}{0.31\textwidth}
        \centering
        \includegraphics[width=\textwidth]{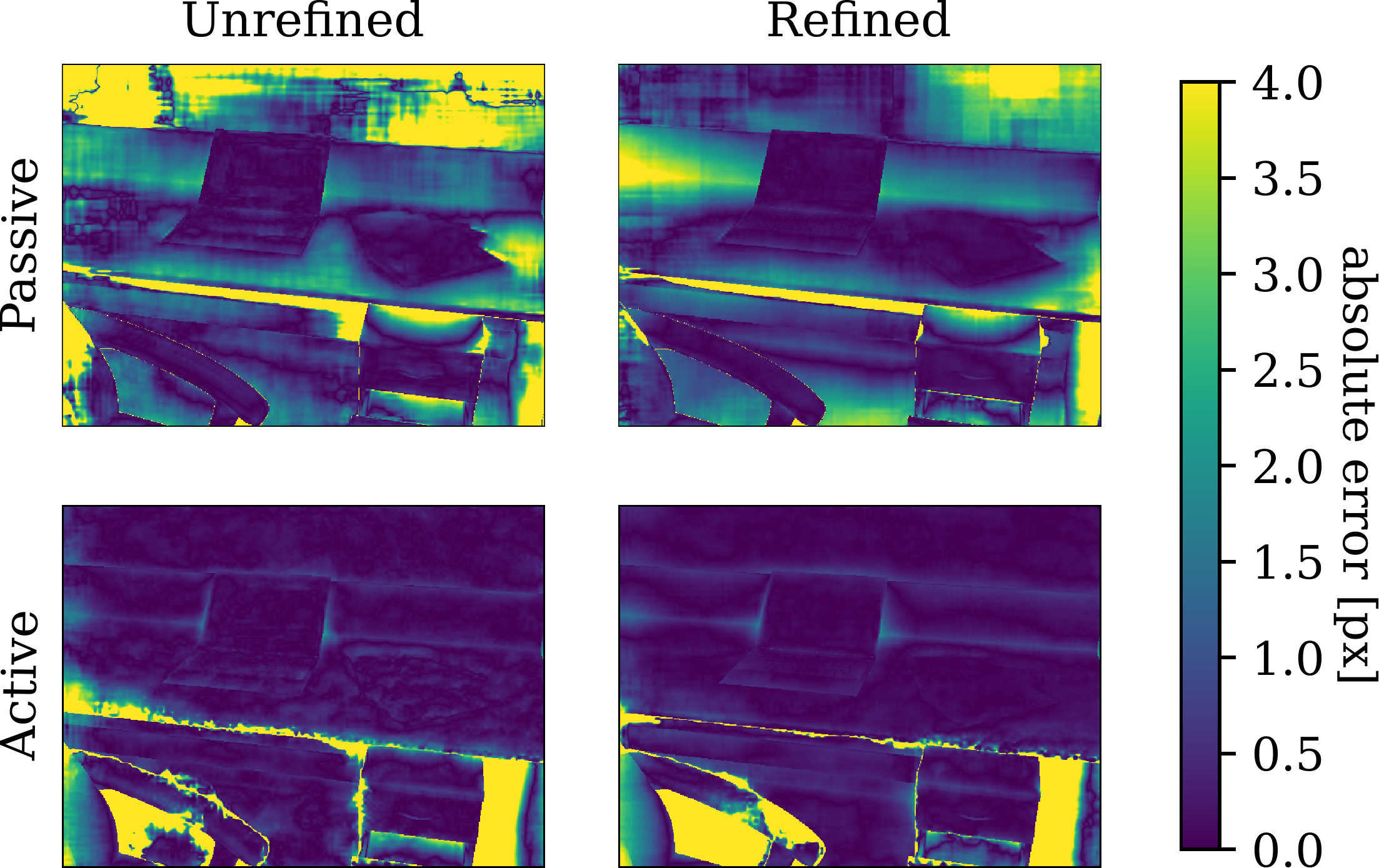}
        \caption{Office of Fig.~\ref{fig:test_set}e with ACVNet \cite{xu2022attention}.}
        \label{fig:errormaps:edges}
    \end{subfigure}
    \hfill
    \begin{subfigure}{0.31\textwidth}
        \centering
        \includegraphics[width=\textwidth]{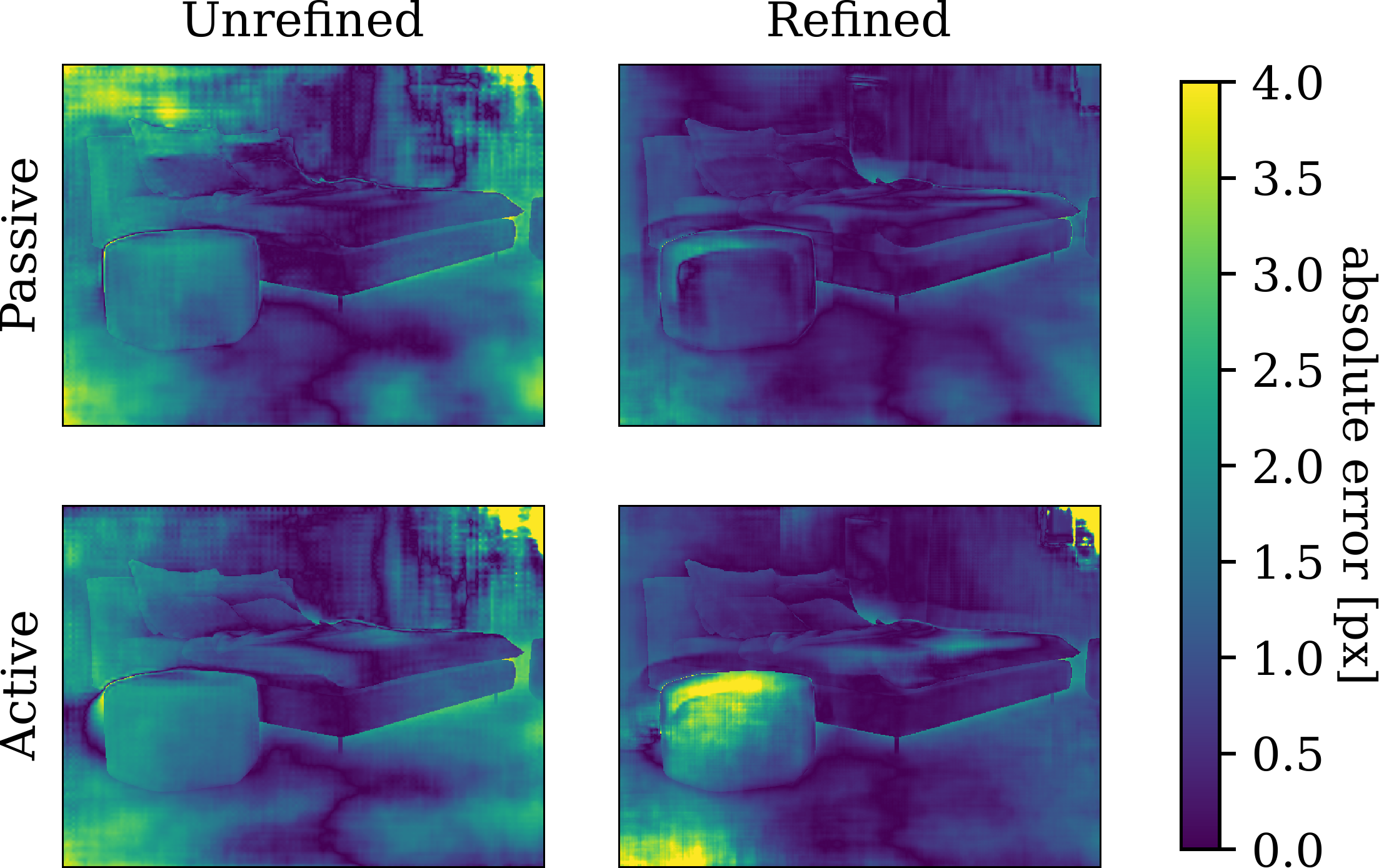}
        \caption{Bedroom of Fig.~\ref{fig:test_set}i with CascadeStereo \cite{gu2019cas}.}
        \label{fig:errormaps:adversarial}
    \end{subfigure}
    \hfill
    \begin{subfigure}{0.31\textwidth}
        \centering
        \includegraphics[width=\textwidth]{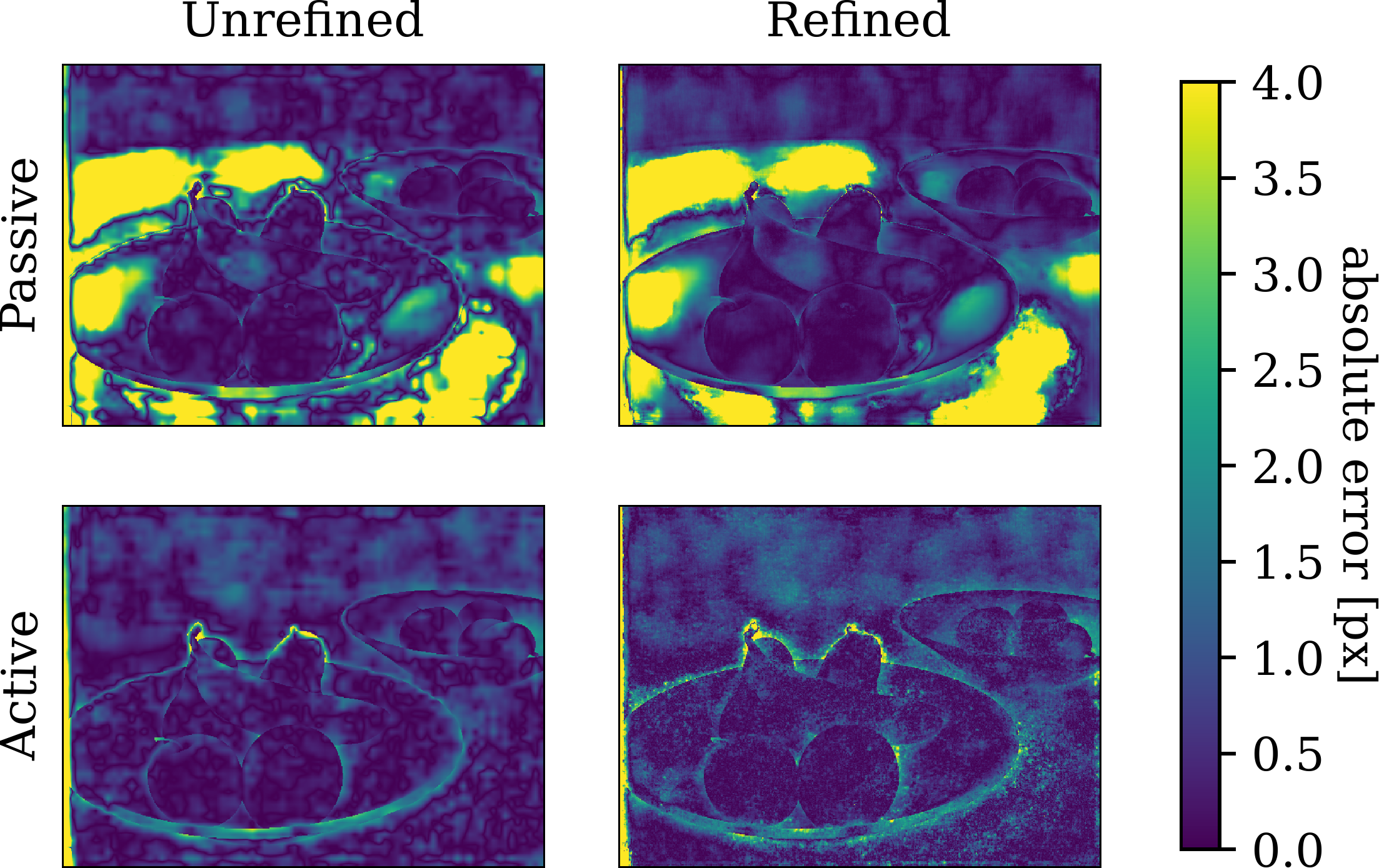}
        \caption{Fruits of Fig.~\ref{fig:test_set}d with StereoNet \cite{khamis2018stereonet}.}
        \label{fig:errormaps:noise}
    \end{subfigure}
    
    \caption{Detailed error maps comparison shows some artifacts not present in the unrefined active disparity appearing in the refined active disparity of certain methods, showing that their refinement module is negatively impacted by the active stereo pseudo random pattern.}
    \label{fig:errormaps}
\end{figure*}

The visual inspection of the results shows that many visual artifacts are difficult to capture by aggregate metrics; see for example Figure~\ref{fig:errormaps}). The refinement module at the end of the stereo matching pipeline is the one most likely to be impacted by a switch from passive to active stereo. Consequently, we chose to test all methods a second time, but with their final refinement module deactivated. Figure~\ref{fig:errormapsclean} shows error maps for methods which did generalize to active stereo while Figure~\ref{fig:errormaps} shows error maps of the methods which had issues generalizing (ACVNet \cite{xu2022attention}, Cascade Stereo  \cite{gu2019cas} and StereoNet \cite{khamis2018stereonet}).

Looking into more details at the error maps of Fig.~\ref{fig:errormaps}, we can notice different effects that may explain these observations. 
In certain rare situations, artifacts will appear in the depth map reconstructed using active stereo; see Fig.~\ref{fig:errormaps:edges} and Fig.~\ref{fig:errormaps:adversarial}. This indicates that under specific circumstances, the pseudo random pattern used for active stereo can act like an adversarial noise. ACVNet has such artifacts around small objects edges, like for example the handle of the chair in Fig.~\ref{fig:errormaps:edges}. ACVNet uses an attention based mechanism to post-process the matching cost volume. This tends to generate edge artifacts in active stereo images. The refinement process is negatively impacted by these artifacts, which further degrade the quality of the final disparity map. 

Cascade Stereo  has some of the most visible examples of such effects, where large error patches seem to appear in the active stereo disparity map, while they are not present in the passive disparity map. If refinement is turned off, the error maps for active and passive stereo are quite similar; see Fig.~\ref{fig:errormaps:adversarial}. Note that the errors between unrefined and refined disparities can vary quite a lot for CascadeStereo compared to the other methods in Figure~\ref{fig:errormapsclean} and Figure~\ref{fig:errormaps}. This shows that CascadeStereo relies extensively on its final refinement module to improve its results, especially compared to other methods. This, in turn, explains why it does not generalize as well as other architectures; see Table~\ref{tab:active_relative_scores}.

StereoNet also exhibits a strange behaviour; see Fig.~\ref{fig:errormaps:noise}. While the network is able to use the pseudo random noise to reconstruct the disparity in uniform regions, the disparity reconstruction appears to be noisy in these areas. Once again, deactivating the refinement module removes the problem. This is not surprising since the StereoNet architecture is an hourglass hierarchical architecture. Other hierarchical methods aggregate their cost volume at different resolution (e.g. \cite{Zhang2019GANet}, \cite{chang2018pyramid}) before making the disparity prediction on the upsampled cost volume. StereoNet makes the prediction on a downsampled cost volume and then uses a module guided only by the initial image to upsample the disparity \cite{khamis2018stereonet}. This approach makes StereoNet fast, but also very reliant on the appearance of the input images. This makes generalization to active stereo difficult. 

\section{Fine-tuning the models which failed to generalize}
\label{sec:finetune}

Can the problems some methods encountered when trying to generalize their predictions from  passive to active stereo image pairs be eliminated by fine-tuning the aforementioned architectures on the active stereo images of our dataset? To test this, we fine-tuned ACVNet \cite{xu2022attention}, Cascade-Stereo \cite{gu2019cas} and StereoNet \cite{khamis2018stereonet} on our test set for $10$ epochs. To avoid any problem of catastrophic forgetting, we ensured that the learning rate is kept low, at $5e-5$ per mini-batch. Each mini-batch is made of four images and each epoch contained $103$ mini-batches. The loss used for training the initial model was used for fine-tuning along with all hyperparameters specified for the original model.

\begin{table}[t]
    \begin{center}
    \caption{Evaluation of the three fine-tuned methods, ACVNet \cite{xu2022attention}, Cascade-Stereo \cite{gu2019cas} and StereoNet \cite{khamis2018stereonet}.
    }
    \label{tab:active_dataset_results_ft}
    \vspace{5pt}
    \scalebox{0.8}{
    \begin{tabular}{lrrrrrr}
        \toprule
        & \multicolumn{6}{c}{Active stereo images} \\
        \cmidrule{2-7}
        Method          & RMSE$_T$ $\downarrow$    & MAE$_T$ $\downarrow$    & BAD$_{0.5}$ $\downarrow$ & BAD$_1$ $\downarrow$  & BAD$_2$ $\downarrow$  & BAD$_4$ $\downarrow$  \\ \midrule
        ACVNet \cite{xu2022attention} (original) & 3.49px & 1.31px & 23\% & 13\% & 8\% & 5\% \\
        ACVNet \cite{xu2022attention} (active stereo fine-tuned) & 2.16px & 0.66px & 17\% & 7\% & 4\% & 3\% \\
        Cascade-Stereo \cite{gu2019cas} (original) & 4.48px & 2.07px & 64\%   & 33\% & 14\% & 8\% \\
        Cascade-Stereo \cite{gu2019cas} (active stereo fine-tuned) &2.11px & 0.66px & 20\%   & 7\% & 4\% & 2\% \\
        StereoNet \cite{khamis2018stereonet} (original) & 7.74px & 1.79px & 44\%   & 23\% & 12\% & 7\%  \\
        StereoNet \cite{khamis2018stereonet} (active stereo fine-tuned) & 7.50px & 1.78px & 56\%   & 29\% & 13\% & 7\%  \\ \bottomrule
    \end{tabular}}
    \end{center}
\end{table}

The results after fine-tuning both methods are reported in Table~\ref{tab:active_dataset_results_ft}. One can observe that ACVNet and Cascade-Stereo exhibit a drastic improvement in their performance.

The visual inspection of the results  reveals that for most image regions, Cascade-Stereo error is now mostly below one pixel, showing no large visible artifacts. ACVNet sees similar improvements to Cascade-Stereo. This shows that the methods were able to adapt to  active stereo after fine-tuning.

For StereoNet, the RMSE improved, the BAD$_{0.5}$ and BAD$_{1}$ metric degraded, while the other metrics stagnated. No more artifacts correlated to the active pattern points distribution are visible. However, new edge artifacts  appear in certain images. This indicates that the fine-tuning is turning the refinement module off instead of adapting it to the projected dot pattern. This, in turn, shows that an architecture largely reliant on the smoothness of the input images is ill-suited for active stereo vision. ActiveStereoNet solved this issue by decoupling the image convolutions from the disparity convolutions in their refinement module \cite{Zhang_2018_ECCV}, which gave it more flexibility to distinguish the active pattern from real objects.

\begin{figure}[t]
    \centering
    
    \begin{subfig}{0.32\textwidth}
        \centering
        \includegraphics[width=\textwidth]{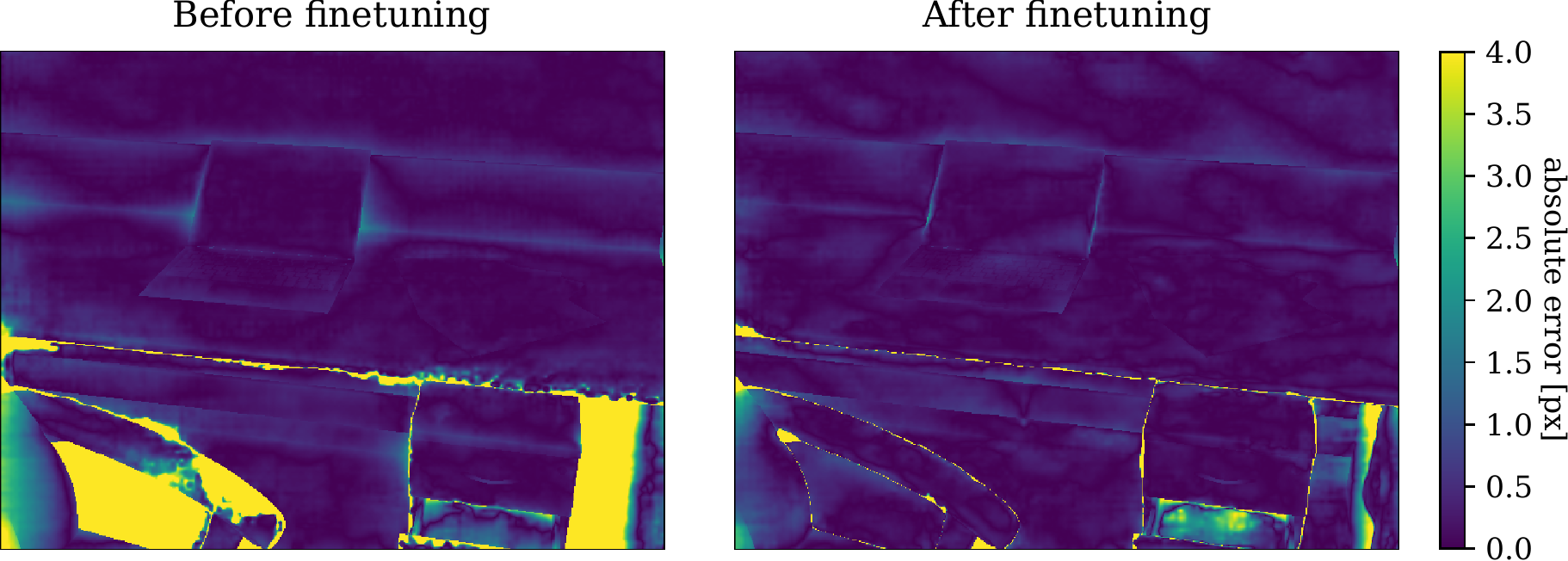}
        \caption{Office of Fig.~\ref{fig:test_set}e with ACVNet \cite{xu2022attention}.}
        \label{fig:results_ft:acvnet}
    \end{subfig}
    \hfill
    \begin{subfig}{0.32\textwidth}
        \centering
        \includegraphics[width=\textwidth]{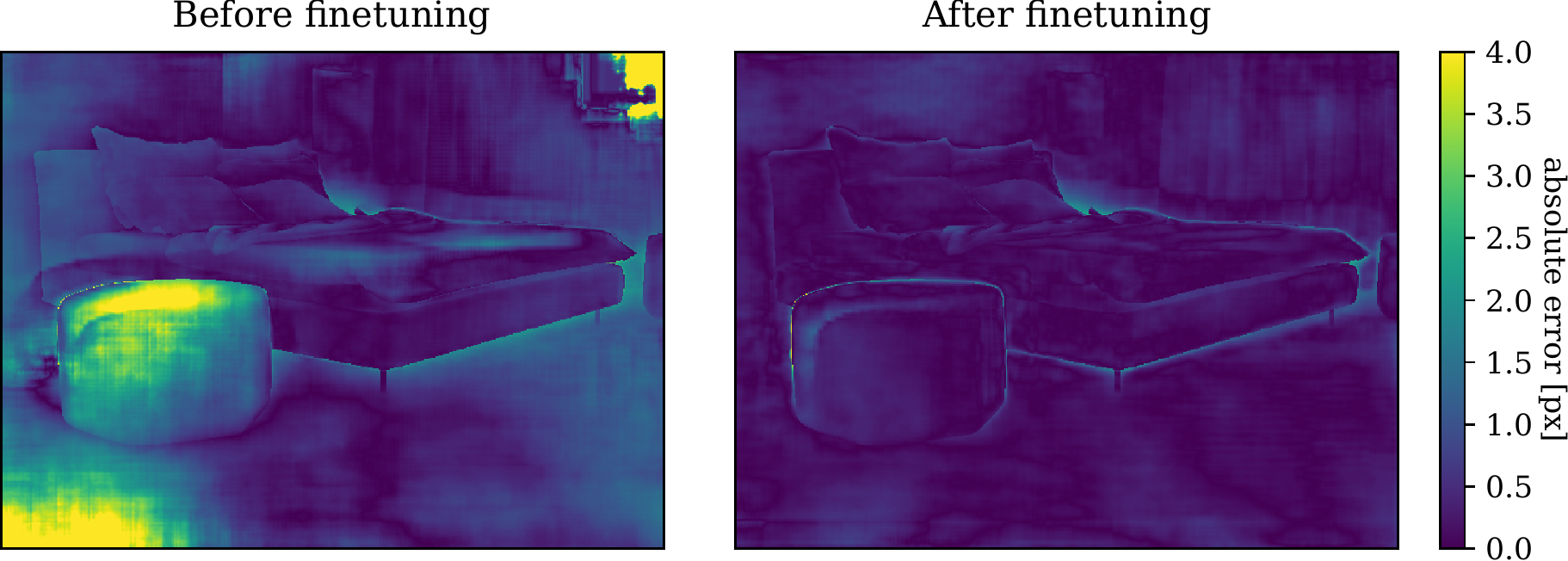}
        \caption{Bedroom of Fig.~\ref{fig:test_set}i with CascadeStereo \cite{gu2019cas}.}
        \label{fig:results_ft:cascadestereo}
    \end{subfig}
    \hfill
    \begin{subfig}{0.32\textwidth}
        \centering
        \includegraphics[width=\textwidth]{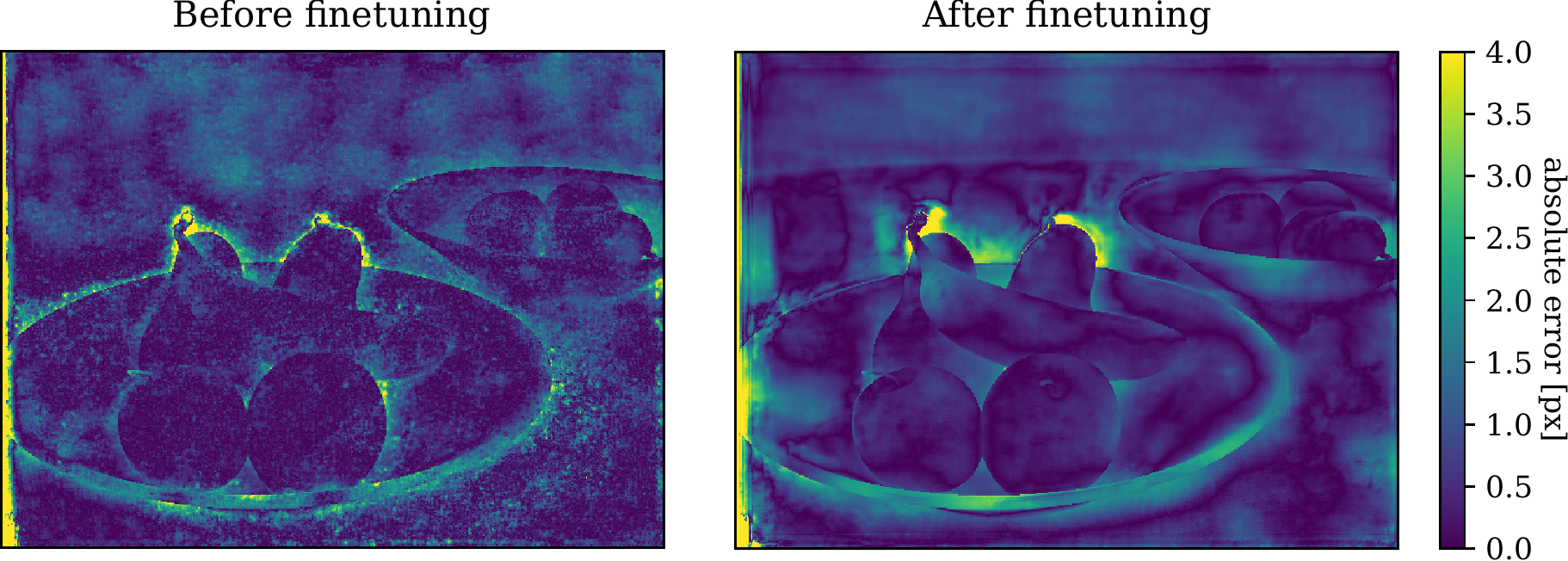}
        \caption{Fruits of Fig.~\ref{fig:test_set}d  with StereoNet \cite{khamis2018stereonet}.}
        \label{fig:results_ft:stereonet}
    \end{subfig}
    
    \caption{Error maps of ACVNet, CascadeStereo and StereoNet when fine-tuned on our dataset.}
    \label{fig:results_ft}
    \vspace{-5pt}
\end{figure}

\section{Conclusion and Perspectives}
\label{sec:conclusion}
We proposed the first dataset and associated benchmark that enables the comparison of the relative performance of stereo vision algorithms when applied to active and passive stereo. 
Using this dataset, we undertook extensive experiments to evaluate the performance of twenty state-of-the-art end-to-end deep learning models. The reported results show that it is possible, to a certain extent, to use methods trained for passive stereo for active stereo vision.
This work also shows that the weak point of those architectures is the final refinement layers. Using our training set, we were able to improve the performances of StereoNet \cite{khamis2018stereonet} and CascadeStereo \cite{gu2019cas}, which had difficulty generalizing to active stereo. StereoNet, the architecture reliant on the smoothness of the input image pair for refinement, still had very poor results, indicating that models that favor shapes prior over appearance priors are more robust. 

Active stereo is an important subfield of stereo-vision. By using our proposed dataset, we were able to examine the generalization ability of current deep learning models. The dataset can also be used to fine-tune these deep learning stereo models for active stereo vision. In the future, we expect to see a growing number of ever larger deep neural networks for stereo vision. Thus, being able to evaluate their generalization ability will become more and more important, and our dataset will prove invaluable in this regard.

\bibliographystyle{plainnat}
\bibliography{bibliography}

\begin{thebibliography}{45}
\providecommand{\natexlab}[1]{#1}
\providecommand{\url}[1]{\texttt{#1}}
\expandafter\ifx\csname urlstyle\endcsname\relax
  \providecommand{\doi}[1]{doi: #1}\else
  \providecommand{\doi}{doi: \begingroup \urlstyle{rm}\Url}\fi

\bibitem[Chang and Chen(2018)]{chang2018pyramid}
Jia-Ren Chang and Yong-Sheng Chen.
\newblock Pyramid stereo matching network.
\newblock In \emph{Proceedings of the IEEE Conference on Computer Vision and
  Pattern Recognition}, pages 5410--5418, 2018.

\bibitem[Chang et~al.(2020)Chang, Chang, and Chen]{Chang_2020_ACCV}
Jia-Ren Chang, Pei-Chun Chang, and Yong-Sheng Chen.
\newblock Attention-aware feature aggregation for real-time stereo matching on
  edge devices.
\newblock In \emph{Proceedings of the Asian Conference on Computer Vision
  (ACCV)}, November 2020.

\bibitem[Chen et~al.(2011)Chen, Li, and Kwok]{doi:10.1177/0278364911410755}
Shengyong Chen, Youfu Li, and Ngai~Ming Kwok.
\newblock Active vision in robotic systems: A survey of recent developments.
\newblock \emph{The International Journal of Robotics Research}, 30\penalty0
  (11):\penalty0 1343--1377, 2011.

\bibitem[Community(2018)]{BlenderSoftware}
Blender~Online Community.
\newblock \emph{Blender - a 3D modelling and rendering package}.
\newblock Blender Foundation, Stichting Blender Foundation, Amsterdam, 2018.
\newblock URL \url{http://www.blender.org}.

\bibitem[Du et~al.(2021)Du, Li, Sun, Zhu, and Tombari]{Hongzhi2021SRH}
Hongzhi Du, Yanyan Li, Yanbiao Sun, Jigui Zhu, and Federico Tombari.
\newblock Srh-net: Stacked recurrent hourglass network for stereo matching.
\newblock \emph{IEEE Robotics and Automation Letters}, 6\penalty0 (4):\penalty0
  8005--8012, 2021.

\bibitem[Duggal et~al.(2019)Duggal, Wang, Ma, Hu, and Urtasun]{9009458}
Shivam Duggal, Shenlong Wang, Wei-Chiu Ma, Rui Hu, and Raquel Urtasun.
\newblock Deeppruner: Learning efficient stereo matching via differentiable
  patchmatch.
\newblock In \emph{2019 IEEE/CVF International Conference on Computer Vision
  (ICCV)}, pages 4383--4392, 2019.

\bibitem[Geiger et~al.(2012)Geiger, Lenz, and Urtasun]{Geiger2012CVPR}
Andreas Geiger, Philip Lenz, and Raquel Urtasun.
\newblock Are we ready for autonomous driving? the kitti vision benchmark
  suite.
\newblock In \emph{Conference on Computer Vision and Pattern Recognition
  (CVPR)}, 2012.

\bibitem[Giralt(2017)]{10.5406/jfilmvideo.69.1.0003}
Gabriel~F. Giralt.
\newblock The interchangeability of vfx and live action and its implications
  for realism.
\newblock \emph{Journal of Film and Video}, 69\penalty0 (1):\penalty0 3--17,
  2017.
\newblock ISSN 07424671, 19346018.

\bibitem[Gu et~al.(2020)Gu, Fan, Zhu, Dai, Tan, and Tan]{gu2019cas}
Xiaodong Gu, Zhiwen Fan, Siyu Zhu, Zuozhuo Dai, Feitong Tan, and Ping Tan.
\newblock Cascade cost volume for high-resolution multi-view stereo and stereo
  matching.
\newblock In \emph{Proceedings of the IEEE/CVF Conference on Computer Vision
  and Pattern Recognition (CVPR)}, June 2020.

\bibitem[Guo et~al.(2019)Guo, Yang, Yang, Wang, and Li]{guo2019group}
Xiaoyang Guo, Kai Yang, Wukui Yang, Xiaogang Wang, and Hongsheng Li.
\newblock Group-wise correlation stereo network.
\newblock In \emph{Proceedings of the IEEE Conference on Computer Vision and
  Pattern Recognition}, pages 3273--3282, 2019.

\bibitem[Hirschmuller and Scharstein(2007)]{hirschmuller2007evaluation}
Heiko Hirschmuller and Daniel Scharstein.
\newblock Evaluation of cost functions for stereo matching.
\newblock In \emph{2007 IEEE Conference on Computer Vision and Pattern
  Recognition}, pages 1--8. IEEE, 2007.

\bibitem[Huang et~al.(2020)Huang, Wang, Cheng, Zhou, Geng, and Yang]{8753527}
Xinyu Huang, Peng Wang, Xinjing Cheng, Dingfu Zhou, Qichuan Geng, and Ruigang
  Yang.
\newblock The apolloscape open dataset for autonomous driving and its
  application.
\newblock \emph{IEEE Transactions on Pattern Analysis and Machine
  Intelligence}, 42\penalty0 (10):\penalty0 2702--2719, 2020.

\bibitem[Keselman et~al.(2017)Keselman, Iselin~Woodfill, Grunnet-Jepsen, and
  Bhowmik]{Keselman_2017_CVPR_Workshops}
Leonid Keselman, John Iselin~Woodfill, Anders Grunnet-Jepsen, and Achintya
  Bhowmik.
\newblock Intel realsense stereoscopic depth cameras.
\newblock In \emph{Proceedings of the IEEE Conference on Computer Vision and
  Pattern Recognition (CVPR) Workshops}, July 2017.

\bibitem[Khamis et~al.(2018)Khamis, Fanello, Rhemann, Kowdle, Valentin, and
  Izadi]{khamis2018stereonet}
Sameh Khamis, Sean Fanello, Christoph Rhemann, Adarsh Kowdle, Julien Valentin,
  and Shahram Izadi.
\newblock Stereonet: Guided hierarchical refinement for real-time edge-aware
  depth prediction.
\newblock In \emph{Proceedings of the European Conference on Computer Vision
  (ECCV), Munich, Germany}, pages 8--14, 2018.

\bibitem[Laga et~al.(2020)Laga, Jospin, Boussaid, and
  Bennamoun]{laga2020survey}
Hamid Laga, Laurent~Valentin Jospin, Farid Boussaid, and Mohammed Bennamoun.
\newblock A survey on deep learning techniques for stereo-based depth
  estimation.
\newblock \emph{IEEE Transactions on Pattern Analysis and Machine
  Intelligence}, 2020.

\bibitem[Levoy et~al.(2005)Levoy, Gerth, Curless, and Pull]{levoy2005stanford}
Marc Levoy, J~Gerth, B~Curless, and K~Pull.
\newblock The stanford 3d scanning repository.
\newblock \emph{URL http://www-graphics. stanford. edu/data/3dscanrep},
  5\penalty0 (10), 2005.

\bibitem[Li et~al.(2022)Li, Wang, Xiong, Cai, Yan, Yang, Liu, Fan, and
  Liu]{li2022practical}
Jiankun Li, Peisen Wang, Pengfei Xiong, Tao Cai, Ziwei Yan, Lei Yang, Jiangyu
  Liu, Haoqiang Fan, and Shuaicheng Liu.
\newblock Practical stereo matching via cascaded recurrent network with
  adaptive correlation.
\newblock In \emph{Proceedings of the IEEE/CVF Conference on Computer Vision
  and Pattern Recognition}, pages 16263--16272, 2022.

\bibitem[Lipson et~al.(2021)Lipson, Teed, and Deng]{lipson2021raft}
Lahav Lipson, Zachary Teed, and Jia Deng.
\newblock Raft-stereo: Multilevel recurrent field transforms for stereo
  matching.
\newblock In \emph{International Conference on 3D Vision (3DV)}, 2021.

\bibitem[Liu et~al.(2022)Liu, Yu, and Long]{Liu_Yu_Long_2022}
Biyang Liu, Huimin Yu, and Yangqi Long.
\newblock Local similarity pattern and cost self-reassembling for deep stereo
  matching networks.
\newblock \emph{Proceedings of the AAAI Conference on Artificial Intelligence},
  36\penalty0 (2):\penalty0 1647--1655, Jun. 2022.

\bibitem[Mayer et~al.(2016)Mayer, Ilg, Hausser, Fischer, Cremers, Dosovitskiy,
  and Brox]{mayer2016large}
Nikolaus Mayer, Eddy Ilg, Philip Hausser, Philipp Fischer, Daniel Cremers,
  Alexey Dosovitskiy, and Thomas Brox.
\newblock A large dataset to train convolutional networks for disparity,
  optical flow, and scene flow estimation.
\newblock In \emph{Proceedings of the IEEE conference on computer vision and
  pattern recognition}, pages 4040--4048, 2016.

\bibitem[Menze and Geiger(2015)]{Menze2015CVPR}
Moritz Menze and Andreas Geiger.
\newblock Object scene flow for autonomous vehicles.
\newblock In \emph{Conference on Computer Vision and Pattern Recognition
  (CVPR)}, 2015.

\bibitem[Nishihara(1984)]{10.1117/12.7973334}
H.~K. Nishihara.
\newblock {Practical Real-Time Imaging Stereo Matcher}.
\newblock \emph{Optical Engineering}, 23\penalty0 (5):\penalty0 536 -- 545,
  1984.

\bibitem[Orts-Escolano et~al.(2016)Orts-Escolano, Rhemann, Fanello, Chang,
  Kowdle, Degtyarev, Kim, Davidson, Khamis, Dou, Tankovich, Loop, Cai, Chou,
  Mennicken, Valentin, Pradeep, Wang, Kang, Kohli, Lutchyn, Keskin, and
  Izadi]{holoportation2016}
Sergio Orts-Escolano, Christoph Rhemann, Sean Fanello, Wayne Chang, Adarsh
  Kowdle, Yury Degtyarev, David Kim, Philip~L. Davidson, Sameh Khamis, Mingsong
  Dou, Vladimir Tankovich, Charles Loop, Qin Cai, Philip~A. Chou, Sarah
  Mennicken, Julien Valentin, Vivek Pradeep, Shenlong Wang, Sing~Bing Kang,
  Pushmeet Kohli, Yuliya Lutchyn, Cem Keskin, and Shahram Izadi.
\newblock Holoportation: Virtual 3d teleportation in real-time.
\newblock In \emph{Proceedings of the 29th Annual Symposium on User Interface
  Software and Technology}, UIST '16, page 741–754, New York, NY, USA, 2016.
  Association for Computing Machinery.
\newblock ISBN 9781450341899.

\bibitem[Paullada et~al.(2021)Paullada, Raji, Bender, Denton, and
  Hanna]{PAULLADA2021100336}
Amandalynne Paullada, Inioluwa~Deborah Raji, Emily~M. Bender, Emily Denton, and
  Alex Hanna.
\newblock Data and its (dis)contents: A survey of dataset development and use
  in machine learning research.
\newblock \emph{Patterns}, 2\penalty0 (11):\penalty0 100336, 2021.
\newblock ISSN 2666--3899.

\bibitem[Pharr et~al.(2016)Pharr, Jakob, and Humphreys]{pharr2016physically}
Matt Pharr, Wenzel Jakob, and Greg Humphreys.
\newblock \emph{Physically based rendering: From theory to implementation}.
\newblock Morgan Kaufmann, 2016.

\bibitem[Planche et~al.(2017)Planche, Wu, Ma, Sun, Kluckner, Lehmann, Chen,
  Hutter, Zakharov, Kosch, and Ernst]{8374552}
Benjamin Planche, Ziyan Wu, Kai Ma, Shanhui Sun, Stefan Kluckner, Oliver
  Lehmann, Terrence Chen, Andreas Hutter, Sergey Zakharov, Harald Kosch, and
  Jan Ernst.
\newblock Depthsynth: Real-time realistic synthetic data generation from cad
  models for 2.5d recognition.
\newblock In \emph{2017 International Conference on 3D Vision (3DV)}, pages
  1--10, 2017.

\bibitem[Pribani{\'{c}} et~al.(2016)Pribani{\'{c}}, Petkovi{\'{c}},
  Djonli{\'{c}}, Angladon, and Gasparini]{10.1007/978-3-319-41501-7_50}
Tomislav Pribani{\'{c}}, Tomislav Petkovi{\'{c}}, Matea Djonli{\'{c}}, Vincent
  Angladon, and Simone Gasparini.
\newblock 3d structured light scanner on the smartphone.
\newblock In Aur{\'e}lio Campilho and Fakhri Karray, editors, \emph{Image
  Analysis and Recognition}, pages 443--450, Cham, 2016. Springer International
  Publishing.
\newblock ISBN 978-3-319-41501-7.

\bibitem[Riegler et~al.(2019)Riegler, Liao, Donne, Koltun, and
  Geiger]{Riegler_2019_CVPR}
Gernot Riegler, Yiyi Liao, Simon Donne, Vladlen Koltun, and Andreas Geiger.
\newblock Connecting the dots: Learning representations for active monocular
  depth estimation.
\newblock In \emph{Proceedings of the IEEE/CVF Conference on Computer Vision
  and Pattern Recognition (CVPR)}, June 2019.

\bibitem[Ryan~Fanello et~al.(2017)Ryan~Fanello, Valentin, Rhemann, Kowdle,
  Tankovich, Davidson, and Izadi]{Fanello_2017_CVPR}
Sean Ryan~Fanello, Julien Valentin, Christoph Rhemann, Adarsh Kowdle, Vladimir
  Tankovich, Philip Davidson, and Shahram Izadi.
\newblock Ultrastereo: Efficient learning-based matching for active stereo
  systems.
\newblock In \emph{Proceedings of the IEEE Conference on Computer Vision and
  Pattern Recognition (CVPR)}, July 2017.

\bibitem[Scharstein and Szeliski(2002)]{scharstein2002taxonomy}
Daniel Scharstein and Richard Szeliski.
\newblock A taxonomy and evaluation of dense two-frame stereo correspondence
  algorithms.
\newblock \emph{International journal of computer vision}, 47\penalty0
  (1):\penalty0 7--42, 2002.

\bibitem[Scharstein et~al.(2014)Scharstein, Hirschm{\"u}ller, Kitajima,
  Krathwohl, Ne{\v{s}}i{\'c}, Wang, and Westling]{scharstein2014high}
Daniel Scharstein, Heiko Hirschm{\"u}ller, York Kitajima, Greg Krathwohl, Nera
  Ne{\v{s}}i{\'c}, Xi~Wang, and Porter Westling.
\newblock High-resolution stereo datasets with subpixel-accurate ground truth.
\newblock In \emph{German conference on pattern recognition}, pages 31--42.
  Springer, 2014.

\bibitem[Shamsafar et~al.(2022)Shamsafar, Woerz, Rahim, and
  Zell]{shamsafar2022mobilestereonet}
Faranak Shamsafar, Samuel Woerz, Rafia Rahim, and Andreas Zell.
\newblock Mobilestereonet: Towards lightweight deep networks for stereo
  matching.
\newblock In \emph{Proceedings of the IEEE/CVF Winter Conference on
  Applications of Computer Vision}, pages 2417--2426, 2022.

\bibitem[Szegedy et~al.(2013)Szegedy, Zaremba, Sutskever, Bruna, Erhan,
  Goodfellow, and Fergus]{szegedy2013intriguing}
Christian Szegedy, Wojciech Zaremba, Ilya Sutskever, Joan Bruna, Dumitru Erhan,
  Ian Goodfellow, and Rob Fergus.
\newblock Intriguing properties of neural networks.
\newblock \emph{arXiv preprint arXiv:1312.6199}, 2013.

\bibitem[Tonioni et~al.(2017)Tonioni, Poggi, Mattoccia, and
  Di~Stefano]{Tonioni_2017_ICCV}
Alessio Tonioni, Matteo Poggi, Stefano Mattoccia, and Luigi Di~Stefano.
\newblock Unsupervised adaptation for deep stereo.
\newblock In \emph{Proceedings of the IEEE International Conference on Computer
  Vision (ICCV)}, Oct 2017.

\bibitem[Tosi et~al.(2021)Tosi, Liao, Schmitt, and Geiger]{Tosi2021CVPR}
Fabio Tosi, Yiyi Liao, Carolin Schmitt, and Andreas Geiger.
\newblock Smd-nets: Stereo mixture density networks.
\newblock In \emph{Conference on Computer Vision and Pattern Recognition
  (CVPR)}, 2021.

\bibitem[Wang et~al.(2019)Wang, Lai, Huang, Wang, van~der Maaten, Campbell, and
  Weinberger]{wang2018anytime}
Yan Wang, Zihang Lai, Gao Huang, Brian~H. Wang, Laurens van~der Maaten, Mark
  Campbell, and Kilian~Q. Weinberger.
\newblock Anytime stereo image depth estimation on mobile devices.
\newblock In \emph{2019 International Conference on Robotics and Automation
  (ICRA)}, pages 5893--5900, 2019.

\bibitem[Worley(1996)]{worley1996cellular}
Steven Worley.
\newblock A cellular texture basis function.
\newblock In \emph{Proceedings of the 23rd annual conference on Computer
  graphics and interactive techniques}, pages 291--294, 1996.

\bibitem[Xu et~al.(2022)Xu, Cheng, Guo, and Yang]{xu2022attention}
Gangwei Xu, Junda Cheng, Peng Guo, and Xin Yang.
\newblock Attention concatenation volume for accurate and efficient stereo
  matching.
\newblock In \emph{Proceedings of the IEEE/CVF Conference on Computer Vision
  and Pattern Recognition}, pages 12981--12990, 2022.

\bibitem[Xu and Zhang(2020)]{xu2020aanet}
Haofei Xu and Juyong Zhang.
\newblock Aanet: Adaptive aggregation network for efficient stereo matching.
\newblock In \emph{Proceedings of the IEEE/CVF Conference on Computer Vision
  and Pattern Recognition}, pages 1959--1968, 2020.

\bibitem[Yang et~al.(2019{\natexlab{a}})Yang, Manela, Happold, and
  Ramanan]{yang2019hsm}
Gengshan Yang, Joshua Manela, Michael Happold, and Deva Ramanan.
\newblock Hierarchical deep stereo matching on high-resolution images.
\newblock In \emph{The IEEE Conference on Computer Vision and Pattern
  Recognition (CVPR)}, June 2019{\natexlab{a}}.

\bibitem[Yang et~al.(2019{\natexlab{b}})Yang, Song, Huang, Deng, Shi, and
  Zhou]{Yang_2019_CVPR}
Guorun Yang, Xiao Song, Chaoqin Huang, Zhidong Deng, Jianping Shi, and Bolei
  Zhou.
\newblock Drivingstereo: A large-scale dataset for stereo matching in
  autonomous driving scenarios.
\newblock In \emph{Proceedings of the IEEE/CVF Conference on Computer Vision
  and Pattern Recognition (CVPR)}, June 2019{\natexlab{b}}.

\bibitem[Zeng et~al.(2021)Zeng, Wang, Zhou, Sun, Huang, Zhang, and
  Wang]{9360804}
Haojie Zeng, Bin Wang, Xiaoping Zhou, Xiaojing Sun, Longxiang Huang, Qian
  Zhang, and Yang Wang.
\newblock Tsfe-net: Two-stream feature extraction networks for active stereo
  matching.
\newblock \emph{IEEE Access}, 9:\penalty0 33954--33962, 2021.

\bibitem[Zhang et~al.(2019)Zhang, Prisacariu, Yang, and Torr]{Zhang2019GANet}
Feihu Zhang, Victor Prisacariu, Ruigang Yang, and Philip~HS Torr.
\newblock Ga-net: Guided aggregation net for end-to-end stereo matching.
\newblock In \emph{Proceedings of the IEEE Conference on Computer Vision and
  Pattern Recognition}, pages 185--194, 2019.

\bibitem[Zhang et~al.(2020)Zhang, Qi, Yang, Prisacariu, Wah, and
  Torr]{zhang2019domaininvariant}
Feihu Zhang, Xiaojuan Qi, Ruigang Yang, Victor Prisacariu, Benjamin Wah, and
  Philip Torr.
\newblock Domain-invariant stereo matching networks.
\newblock In \emph{Europe Conference on Computer Vision (ECCV)}, 2020.

\bibitem[Zhang et~al.(2018)Zhang, Khamis, Rhemann, Valentin, Kowdle, Tankovich,
  Schoenberg, Izadi, Funkhouser, and Fanello]{Zhang_2018_ECCV}
Yinda Zhang, Sameh Khamis, Christoph Rhemann, Julien Valentin, Adarsh Kowdle,
  Vladimir Tankovich, Michael Schoenberg, Shahram Izadi, Thomas Funkhouser, and
  Sean Fanello.
\newblock Activestereonet: End-to-end self-supervised learning for active
  stereo systems.
\newblock In \emph{Proceedings of the European Conference on Computer Vision
  (ECCV)}, September 2018.

\end{thebibliography}


\begin{thebibliography}{22}
\providecommand{\natexlab}[1]{#1}
\providecommand{\url}[1]{\texttt{#1}}
\expandafter\ifx\csname urlstyle\endcsname\relax
  \providecommand{\doi}[1]{doi: #1}\else
  \providecommand{\doi}{doi: \begingroup \urlstyle{rm}\Url}\fi

\bibitem[ble({\natexlab{a}})]{blendermarket}
Blender market.
\newblock \url{https://blendermarket.com/}, {\natexlab{a}}.
\newblock Accessed: 2022-05-31.

\bibitem[ble({\natexlab{b}})]{blendswap}
Blendswap.
\newblock \url{https://www.blendswap.com/}, {\natexlab{b}}.
\newblock Accessed: 2022-05-31.

\bibitem[exr()]{exrtools}
Exr tools.
\newblock \url{https://github.com/french-paragon/exr-tools}.
\newblock Accessed: 2022-05-31.

\bibitem[iee()]{ieeedataport}
{IEEE} dataport.
\newblock \url{https://ieee-dataport.org}.
\newblock Accessed: 2022-05-31.

\bibitem[oci()]{ocio}
Opencolorio.
\newblock \url{https://opencolorio.org/}.
\newblock Accessed: 2022-05-31.

\bibitem[ope()]{openexr}
Openexr.
\newblock \url{https://www.openexr.com/}.
\newblock Accessed: 2022-05-31.

\bibitem[Carroll(2006)]{carroll2006creative}
Michael~W Carroll.
\newblock Creative commons and the new intermediaries.
\newblock \emph{Mich. St. L. Rev.}, page~45, 2006.

\bibitem[Community(2018)]{BlenderSoftware}
Blender~Online Community.
\newblock \emph{Blender - a 3D modelling and rendering package}.
\newblock Blender Foundation, Stichting Blender Foundation, Amsterdam, 2018.
\newblock URL \url{http://www.blender.org}.

\bibitem[Gu et~al.(2020)Gu, Fan, Zhu, Dai, Tan, and Tan]{gu2019cas}
Xiaodong Gu, Zhiwen Fan, Siyu Zhu, Zuozhuo Dai, Feitong Tan, and Ping Tan.
\newblock Cascade cost volume for high-resolution multi-view stereo and stereo
  matching.
\newblock In \emph{Proceedings of the IEEE/CVF Conference on Computer Vision
  and Pattern Recognition (CVPR)}, June 2020.

\bibitem[Hirschmuller(2005)]{sgmHirschmuller}
H.~Hirschmuller.
\newblock Accurate and efficient stereo processing by semi-global matching and
  mutual information.
\newblock In \emph{2005 IEEE Computer Society Conference on Computer Vision and
  Pattern Recognition (CVPR'05)}, volume~2, pages 807--814 vol. 2, 2005.

\bibitem[Jospin et~al.(2022)Jospin, Laga, Boussaid, and Bennamoun]{our_dataset}
Laurent~Valentin Jospin, Hamid Laga, Farid Boussaid, and Mohammed Bennamoun.
\newblock Active-passive simstereo, 2022.
\newblock URL \url{https://dx.doi.org/10.21227/gf1e-t452}.

\bibitem[Jospin et~al.(N.D.)Jospin, Laga, Boussaid, and Bennamoun]{stereobnn}
Laurent~Valentin Jospin, Hamid Laga, Farid Boussaid, and Mohammed Bennamoun.
\newblock Bayesian learning for disparity map refinement for semi-dense active
  stereo vision.
\newblock unpublished, N.D.

\bibitem[Kainz et~al.(2009)Kainz, Bogart, and Stanczyk]{kainz2009technical}
Florian Kainz, Rod Bogart, and Piotr Stanczyk.
\newblock Technical introduction to openexr.
\newblock \emph{Industrial light and magic}, 21, 2009.

\bibitem[Keselman et~al.(2017)Keselman, Iselin~Woodfill, Grunnet-Jepsen, and
  Bhowmik]{Keselman_2017_CVPR_Workshops}
Leonid Keselman, John Iselin~Woodfill, Anders Grunnet-Jepsen, and Achintya
  Bhowmik.
\newblock Intel realsense stereoscopic depth cameras.
\newblock In \emph{Proceedings of the IEEE Conference on Computer Vision and
  Pattern Recognition (CVPR) Workshops}, July 2017.

\bibitem[Khamis et~al.(2018)Khamis, Fanello, Rhemann, Kowdle, Valentin, and
  Izadi]{khamis2018stereonet}
Sameh Khamis, Sean Fanello, Christoph Rhemann, Adarsh Kowdle, Julien Valentin,
  and Shahram Izadi.
\newblock Stereonet: Guided hierarchical refinement for real-time edge-aware
  depth prediction.
\newblock In \emph{Proceedings of the European Conference on Computer Vision
  (ECCV), Munich, Germany}, pages 8--14, 2018.

\bibitem[Levoy et~al.(2005)Levoy, Gerth, Curless, and Pull]{levoy2005stanford}
Marc Levoy, J~Gerth, B~Curless, and K~Pull.
\newblock The stanford 3d scanning repository.
\newblock \emph{URL http://www-graphics. stanford. edu/data/3dscanrep},
  5\penalty0 (10), 2005.

\bibitem[Osband(2016)]{osband2016risk}
Ian Osband.
\newblock Risk versus uncertainty in deep learning: Bayes, bootstrap and the
  dangers of dropout.
\newblock In \emph{NIPS workshop on bayesian deep learning}, volume 192, 2016.

\bibitem[Scharstein and Szeliski(2002)]{scharstein2002taxonomy}
Daniel Scharstein and Richard Szeliski.
\newblock A taxonomy and evaluation of dense two-frame stereo correspondence
  algorithms.
\newblock \emph{International journal of computer vision}, 47\penalty0
  (1):\penalty0 7--42, 2002.

\bibitem[Selan(2012)]{10.1145/2343483.2343492}
Jeremy Selan.
\newblock Cinematic color: From your monitor to the big screen.
\newblock In \emph{ACM SIGGRAPH 2012 Courses}. Association for Computing
  Machinery, 2012.
\newblock ISBN 9781450316781.

\bibitem[Walker et~al.(2021)Walker, Payne, Hodoul, and
  Dolan]{10.1145/3450508.3464600}
Doug Walker, Carol Payne, Patrick Hodoul, and Michael Dolan.
\newblock Color management with opencolorio v2.
\newblock In \emph{ACM SIGGRAPH 2021 Courses}, SIGGRAPH '21, New York, NY, USA,
  2021. Association for Computing Machinery.
\newblock ISBN 9781450383615.

\bibitem[Xu et~al.(2022)Xu, Cheng, Guo, and Yang]{xu2022attention}
Gangwei Xu, Junda Cheng, Peng Guo, and Xin Yang.
\newblock Attention concatenation volume for accurate and efficient stereo
  matching.
\newblock In \emph{Proceedings of the IEEE/CVF Conference on Computer Vision
  and Pattern Recognition}, pages 12981--12990, 2022.

\bibitem[Zhang et~al.(2018)Zhang, Khamis, Rhemann, Valentin, Kowdle, Tankovich,
  Schoenberg, Izadi, Funkhouser, and Fanello]{Zhang_2018_ECCV}
Yinda Zhang, Sameh Khamis, Christoph Rhemann, Julien Valentin, Adarsh Kowdle,
  Vladimir Tankovich, Michael Schoenberg, Shahram Izadi, Thomas Funkhouser, and
  Sean Fanello.
\newblock Activestereonet: End-to-end self-supervised learning for active
  stereo systems.
\newblock In \emph{Proceedings of the European Conference on Computer Vision
  (ECCV)}, September 2018.

\end{thebibliography}

\end{document}


\title{\datasetname{} - Supplementary material}

\author{
Laurent~Jospin$^{1*}$ \quad Allen~Antony$^1$ \quad Lian~Xu$^1$ \quad Hamid~Laga$^2$ \quad Farid~Boussaid$^1$ \\ \textbf{Mohammed~Bennamoun}$^1$ \\
$^1$University of Western Australia \quad $^2$Murdoch University\\
\texttt{\{laurent.jospin,lian.xu,farid.boussaid,mohammed.bennamoun\}@uwa.edu.au}\\
\texttt{H.Laga@murdoch.edu.au}
}


\maketitle

\section{Organisation of this document}

This document provides additional information about the \datasetname{}  dataset. Section~\ref{sec:license} confirms and motivates the choice of the license for the dataset. The author statement in case of violation of rights is presented in Section~\ref{sec:right}. Section~\ref{sec:filesformat} lists the file formats used in the dataset, their purpose and how to use them. Section~\ref{sec:nonlearning} evaluate traditional non-learning methods on our dataset. Section~\ref{sec:realdata} evaluate methods fine-tuned on the \datasetname{} dataset on real images. Section~\ref{sec:boostraping} checks whether our dataset is large enough for its intended purposes. Section~\ref{sec:colormanagement} presents the problematic of color management. Section~\ref{sec:simulator} present the simulation toolkit we used to generate the dataset. Finally, Section~\ref{sec:storage} presents the plan for the long-term storage of the dataset.

\section{License}
\label{sec:license}

The \datasetname{} dataset is published under a CC-BY 4.0  creative commons license (\url{https://creativecommons.org/licenses/by/4.0/}). This license is fit for research assets as it grants to the end-user the right to use, modify, and share the data  provided that proper citation to the original work is provided \cite{carroll2006creative}.

The \datasetname{} dataset was generated using CC0 (public domain) assets from the Blendswap \cite{blendswap} website and commercial assets from the Blender Market \cite{blendermarket} website. We also used some public models from the Standford 3D repository \cite{levoy2005stanford} (the Stanford Bunny and Dragon, mostly as easter eggs).

\section{Author statement in case of violation of rights}
\label{sec:right}

The authors of the dataset bear all responsibility in case of a violation of right when the dataset was created.

\section{Files and file formats}
\label{sec:filesformat}

The dataset contains the active and passive image pairs (provided as .jpg and .exr images), as well as the ground truth disparities for each image (provided as .npy, .pfm and .exr files).

\begin{description}
\item[Jpg images:] The jpg image format is a very popular, lossy image format, storing a colored image with $8$ bits per channels. We use this format to provide the color managed images in display refered format (see Section~\ref{sec:colormanagement}). Jpg images can be loaded by any image application or library.

\item[Exr images:] The exr image format \cite{kainz2009technical} is a high dynamic range image format popular in the visual effect industry. It  uses a $16$ bit or $32$ bit floating-point value per channel. An exr image can contain multiple layers, each of which is made of one or multiple channels. This means that a single file can contain both the left and right view for active and passive images, as well as the disparity in floating-point format. Working with exr images can be slightly more tedious than with traditional jpg or png images. Blender \cite{BlenderSoftware}, as well as other image processing or visual effects software such as Photoshop, Krita, Nuke, and Natron, can be used to open and inspect exr images. While some image libraries can open exr images, some do not support the layer system. When programming in C++, we recommend using the official OpenExr library \cite{openexr}. When using  Python, we wrote our own tool to work with exr images, which is available on GitHub under an open-source license \cite{exrtools}.

\item[Npy files:] The npy file format is a format used by the numpy library to save and load multidimensional arrays. It is one of the most convenient file format to pass floating point image data to python. For convenience, we do provide the disparity map in the npy format.

\item[Pfm files.] The pfm file format is an old format to transfer floating points images. Writing a parser for pfm files is straightforward. Also,  many scientific image processing libraries can open pfm files. We  provide the disparity map in the pfm format, for convenience when using lower level programming languages like C or C++.
\end{description}

\section{Testing traditional non-learning methods on the Active-Passive SimStereo dataset}
\label{sec:nonlearning}

Traditional non-learning methods are known to generalize very well on active stereo \cite{Keselman_2017_CVPR_Workshops}. Comparing their relative performances against deep learning methods is important to check whether or not deep learning methods can outperform traditional methods.

Evaluated methods include traditional local matching \cite{scharstein2002taxonomy} and semi-global matching\cite{sgmHirschmuller}. We have used a $9 \times 9$ correlation window and a ZNCC cost function. For the semi-global matching method, we used 8 directions for the cost aggregation, with the parameters $P1$ and $P2$ set to $0.001$ and $0.01$.

\begin{table}[ht]
    \begin{center}
\caption{Evaluation results of traditional non learning methods on passive and then active stereo images.}
    \label{tab:tradi_dataset_results}
    \vspace{5pt}
    \resizebox{\columnwidth}{!}{
    \begin{tabular}{lrrrrrrcrrrrrr}
        \toprule
        & \multicolumn{6}{c}{Passive stereo images} & & \multicolumn{6}{c}{Active stereo images}\\
        \cmidrule{2-7}\cmidrule{9-14}
        Method          & RMSE$_T$ $\downarrow$    & MAE$_T$ $\downarrow$    & BAD$_{0.5}$ $\downarrow$ & BAD$_1$ $\downarrow$  & BAD$_2$ $\downarrow$  & BAD$_4$ $\downarrow$  & & RMSE$_T$ $\downarrow$  & MAE$_T$ $\downarrow$    & BAD$_{0.5}$ $\downarrow$ & BAD$_1$ $\downarrow$ & BAD$_2$ $\downarrow$ & BAD$_4$ $\downarrow$ \\ \midrule
        Local matching \cite{scharstein2002taxonomy} & 28.10px & 11.79px & 49\% & 43\% & 37\% & 32\% & & 24.25px & 7.03px & 20\% & 17\% & 16\% & 14\% \\
        Semis-global matching \cite{sgmHirschmuller} & 23.03px & 8.95px & 45\% & 37\% & 31\% & 26\% & & 22.48px & 6.47px & 20\% & 17\% & 15\% & 14\% \\
        \bottomrule
    \end{tabular}}
    \end{center}
\end{table}

The results, reported in Table~\ref{tab:tradi_dataset_results}, show that traditional methods are unable to outperform a number of the deep learning methods that have been analyzed in the main paper (even before finetuning). For the RMSE and MAE metric, all deep learning methods outperform the traditional ones, but this can be misleading as traditional methods can have very high errors for a few single mismatched pixels. In contrast, deep learning methods tend to smooth out their solutions. For the BAD$_N$ metrics, traditional methods are able to outperform a limited set of older deep learning methods (that have not been finetuned).

This shows that even though it is true that traditional methods perform well on active stereo, it does not mean that deep learning models are not useful for active stereo.

\section{Testing the methods finetuned with the Active-Passive SimStereo dataset on real data}
\label{sec:realdata}

\begin{figure}[tb]
    \centering
    \begin{subfigure}[b]{0.48\textwidth}
    \includegraphics[width=\textwidth]{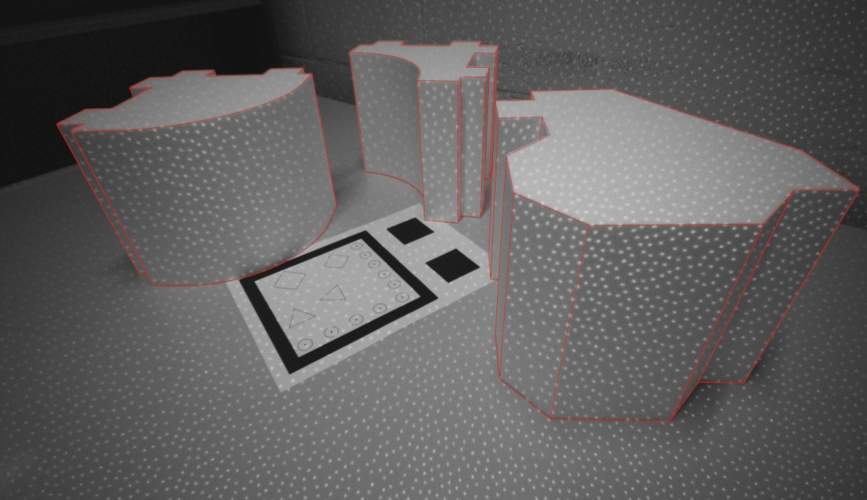}
    \caption{}
    \label{fig:realsenseimgpair:left}
    \end{subfigure}
    \hfill
    \begin{subfigure}[b]{0.48\textwidth}
    \includegraphics[width=\textwidth]{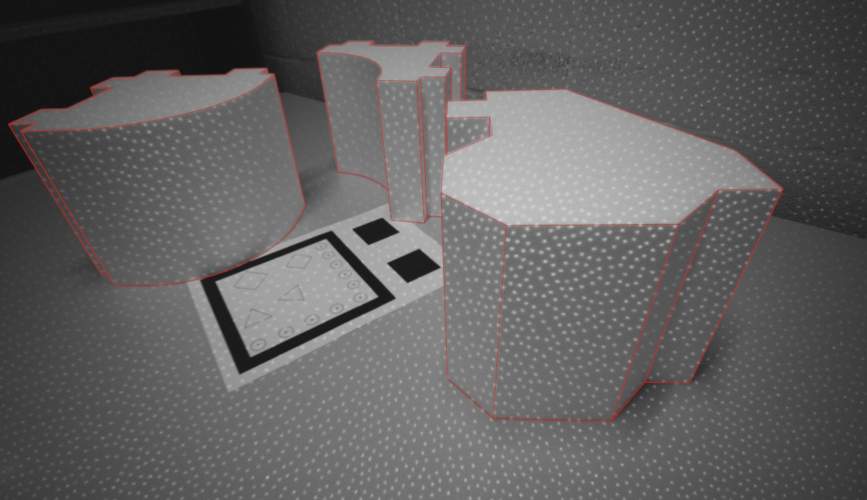}
    \caption{}
    \label{fig:realsenseimgpair:right}
    \end{subfigure}
    
    \caption{Left (\subref{fig:realsenseimgpair:left}) and right  (\subref{fig:realsenseimgpair:right}) frames of an Image pair from the Shapes dataset with the outlined CAD models aligned onto the real scene.}
    \label{fig:realsenseimgpair}
\end{figure}

While our dataset is more aimed at evaluating the generalization abilities of deep learning models across passive and active stereo, it could also be used by the community to train active stereo models. One could thus evaluate whether our dataset can be used to improve the performances of deep learning models on real active stereo images.

To this end, we used the Shapes dataset \cite{stereobnn}, a dataset of active stereo images of geometric shapes cut in polystyrene. The CAD models of the shapes were then re-aligned with the images to generate high quality ground truth disparities; see Figure~\ref{fig:realsenseimgpair}.

Finetuned methods were evaluated on the Shapes dataset, both with their original weights and with the weights obtained after finetuning. We also finetuned and evaluated ActiveStereoNet \cite{Zhang_2018_ECCV}, which was originally trained on active stereo images, but in a self-supervised manner. Since ActiveStereoNet \cite{Zhang_2018_ECCV} and StereoNet \cite{khamis2018stereonet} are related, we used the same loss as StereoNet to finetune ActiveStereoNet in a supervised manner. All other methods were trained, as described in the main paper, by using the same loss and hyperparameters as originally used by the method authors.

\begin{table}[ht]
    \begin{center}
    \caption{Evaluation of Fine-tuned methods on the Shapes dataset.
    }
    \label{tab:shape_dataset_results_ft}
    \vspace{5pt}
    \scalebox{0.8}{
    \begin{tabular}{lrrrrrr}
        \toprule
        & \multicolumn{6}{c}{Shapes active stereo images} \\
        \cmidrule{2-7}
        Method          & RMSE$_T$ $\downarrow$    & MAE$_T$ $\downarrow$    & BAD$_{0.5}$ $\downarrow$ & BAD$_1$ $\downarrow$  & BAD$_2$ $\downarrow$  & BAD$_4$ $\downarrow$  \\ \midrule
        ACVNet \cite{xu2022attention} (original) & 1.86px & 0.91px & 43\% & 20\% & 9\% & 4\% \\
        ACVNet \cite{xu2022attention} (active stereo fine-tuned) & 1.58px & 0.63px & 34\% & 9\% & 3\% & 2\% \\
        Cascade-Stereo \cite{gu2019cas} (original) & 2.55px & 1.47px & 62\%   & 38\% & 18\% & 8\% \\
        Cascade-Stereo \cite{gu2019cas} (active stereo fine-tuned) & 1.61px & 0.69px & 38\%   & 13\% & 5\% & 2\% \\
        StereoNet \cite{khamis2018stereonet} (original) & 4.72px & 2.07px & 64\%   & 40\% & 19\% & 8\%  \\
        StereoNet \cite{khamis2018stereonet} (active stereo fine-tuned) & 4.37px & 1.94px & 70\%   & 48\% & 20\% & 6\%  \\
        ActiveStereoNet \cite{Zhang_2018_ECCV} (original self-supervised) & 8.47px & 2.84px & 62\%   & 39\% & 20\% & 9\%  \\
        ActiveStereoNet \cite{Zhang_2018_ECCV} (active stereo fine-tuned) & 2.04px & 1.07px & 58\%   & 30\% & 11\% & 4\%  \\ \bottomrule
    \end{tabular}}
    \end{center}
\end{table}

The results, reported in Table~\ref{tab:shape_dataset_results_ft}, show clearly that finetuning on our synthetic dataset allowed the different models to improve their performances on real active stereo images, obtained with a RealSense camera. Even ActiveStereoNet \cite{Zhang_2018_ECCV}, which was originally trained for active stereo but in a self supervised manner, see an improvement when trained in a supervised manner on our dataset. StereoNet \cite{khamis2018stereonet} is the only exception, with the RMSE$_T$, MAE$_T$ and BAD$_4$ metrics improving but the BAD$_{0.5}$, BAD$_1$ and BAD$_2$ metrics getting worse. This is due to the fact, as highlithed in the main paper, that the refinement module of StereoNet struggles with active stereo. This, in turn implies that the finetuning made the network focus more on large scale errors.

\section{Impact of the dataset size on evaluating generalization ability}
\label{sec:boostraping}

We are interested in checking if our \datasetname{} dataset is ``sufficiently large". In practice, we say that a dataset is sufficiently large  for the purpose of evaluating the generalization ability of deep learning stereo models if the uncertainty associated with the size of the dataset is negligible compared to the measure of interest. This can be measured by using bootstrapping \cite{osband2016risk}, a method where a subset of a sample is selected at random to estimate the variability of a statistical estimator. This approach allows to get a confidence interval on our conclusions.

First, we want to check if the uncertainty due to the size of the dataset is negligible compared to the relative variations of performances between active and passive stereo. To this end, we randomly sample a subset $T_{50\%}$ of the test set $T$ and recompute the metrics of interest. Here, we work with the $MAE$ and $BAD_2$ metrics, but the code provided with this supplementary material can compute the values for any of the metrics we measured. We denote by $M_{50\%}$ the metric $M$ computed on $T_{50\%}$ instead of $T$. Similarly,  $BAD2_{50\%}$ refers to $BAD_2$ measured on $T_{50\%}$. We then compute the absolute relative variation between each metric $M$ and $M_{50\%}$ defined as:
\begin{equation}
    R(M, M_{50\%}) = \dfrac{|M - M_{50\%}|}{|M|}.
\end{equation}

\noindent As $M_{50\%}$ is a random variable, we need an estimator for $R(M, M_{50\%})$. To this end, we sample $N$ different random $T_{50\%}$ and use these to estimate the expected value $\mathbb{E}[R(M, M_{50\%})]$ of $R(M, M_{50\%})$. For our experiments, we choose $N$ = 50. $\mathbb{E}[R(M, M_{50\%})]$ gives an estimate of the uncertainty of the value of $M_{50\%}$ for a dataset that is half the size of our test set. Using the central limit theorem, the uncertainty for the full test set of the value of $M$ can be estimated by scaling down $\mathbb{E}[R(M, M_{50\%})]$ by a factor of $1/\sqrt{2}$. We finally compare $\mathbb{E}[R(M, M_{50\%})]$ to $R(M_{\text{pas}}, M_{\text{act}})$, the absolute relative variation between the passive and active version of $M$. The results are reported in Table~\ref{tab:boostrap_results}.

\begin{table}[tb]
    \centering
    
    \caption{Expected $MAE$ and $BAD_2$ variations between active and passive images, and within passive or active images}
    \label{tab:boostrap_results}
    \vspace{5pt}
    \resizebox{\columnwidth}{!}{
    \begin{tabular}{lrrrcrrr}
\toprule
{} & \multicolumn{3}{c}{$MAE$} & & \multicolumn{3}{c}{$BAD_2$} \\
\cmidrule{2-4}\cmidrule{6-8}
Method & $R(M_{\text{pas}}, M_{\text{act}})$ & $\mathbb{E}[R(M, M_{50\%})]$ passive & $\mathbb{E}[R(M, M_{50\%})]$ active & & $R(M_{\text{pas}}, M_{\text{act}})$ & $\mathbb{E}[R(M, M_{50\%})]$ passive & $\mathbb{E}[R(M, M_{50\%})]$ active \\
\midrule
AANet & 50\% & 8\% & 5\% & & 59\% & 5\% & 6\% \\
ACVNet & 67\% & 16\% & 12\% & & 56\% & 8\% & 9\% \\
AnyNet & 36\% & 7\% & 6\% & & 30\% & 2\% & 2\% \\
CREStereo & 54\% & 8\% & 9\% & & 73\% & 10\% & 10\% \\
CascadeStereo & 58\% & 11\% & 8\% & &  41\% & 6\% & 7\% \\
Deep-Pruner (best) & 70\% & 13\% & 10\% & & 70\% & 6\% & 9\% \\
Deep-Pruner (fast) & 59\% & 10\% & 9\% & & 59\% & 5\% & 8\% \\
GANet & 61\% & 7\% & 6\% & & 67\% & 5\% & 7\% \\
GwcNet & 65\% & 7\% & 7\% & & 65\% & 4\% & 7\% \\
HighResStereo & 56\% & 12\% & 8\% & & 58\% & 6\% & 9\% \\
Lac-GwcNet & 68\% & 11\% & 14\% & & 69\% & 5\% & 10\% \\
MobileStereoNet2D & 53\% & 6\% & 8\% & & 61\% & 4\% & 7\% \\
MobileStereoNet3D & 63\% & 7\% & 7\% & & 66\% & 3\% & 7\% \\
PSMNet & 45\% & 7\% & 3\% & & 39\% & 3\% & 4\% \\
RAFT-Stereo & 58\% & 11\% & 8\% & & 57\% & 6\% & 8\% \\
RealTimeStereo & 39\% & 6\% & 6\% & & 36\% & 3\% & 3\% \\
SMD-Nets & 65\% & 7\% & 10\% & & 63\% & 4\% & 7\% \\
SRHNet & 62\% & 10\% & 8\% & & 68\% & 4\% & 7\% \\
StereoNet & 56\% & 6\% & 5\% & & 54\% & 3\% & 5\% \\
ActiveStereoNet & 75\% & 6\% & 5\% & & 61\% & 4\% & 4\% \\
\bottomrule

\end{tabular}
    }
\end{table}

\begin{figure}[pt]
    \centering
    
    \begin{subfigure}{\textwidth}
        \centering
        \includegraphics[width=\textwidth]{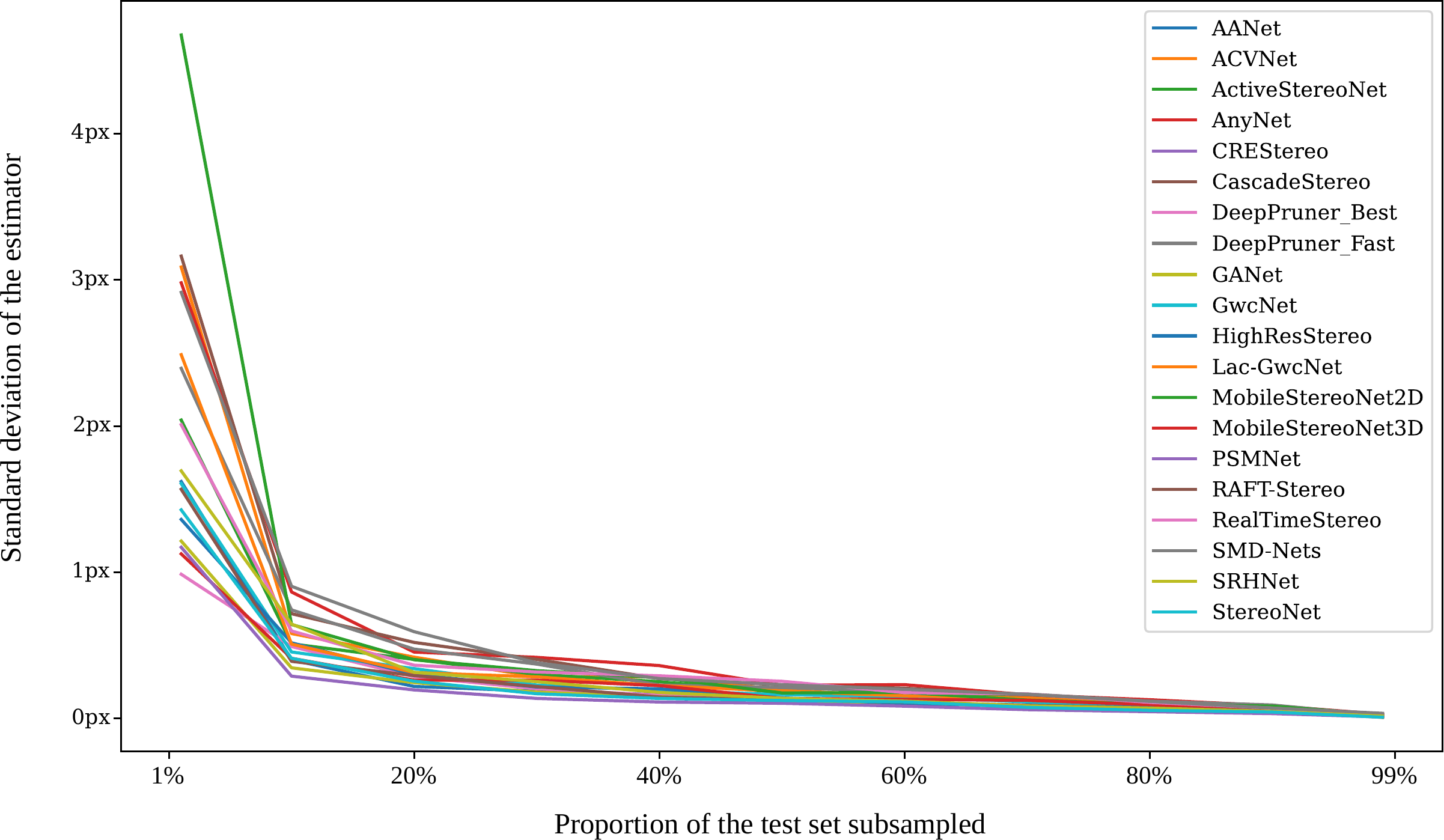}
        \caption{$MAE$}
        \label{fig:boostrap_variance:MAE}
    \end{subfigure}
    
    \vspace{15pt}
    
    \begin{subfigure}{\textwidth}
        \centering
        \includegraphics[width=\textwidth]{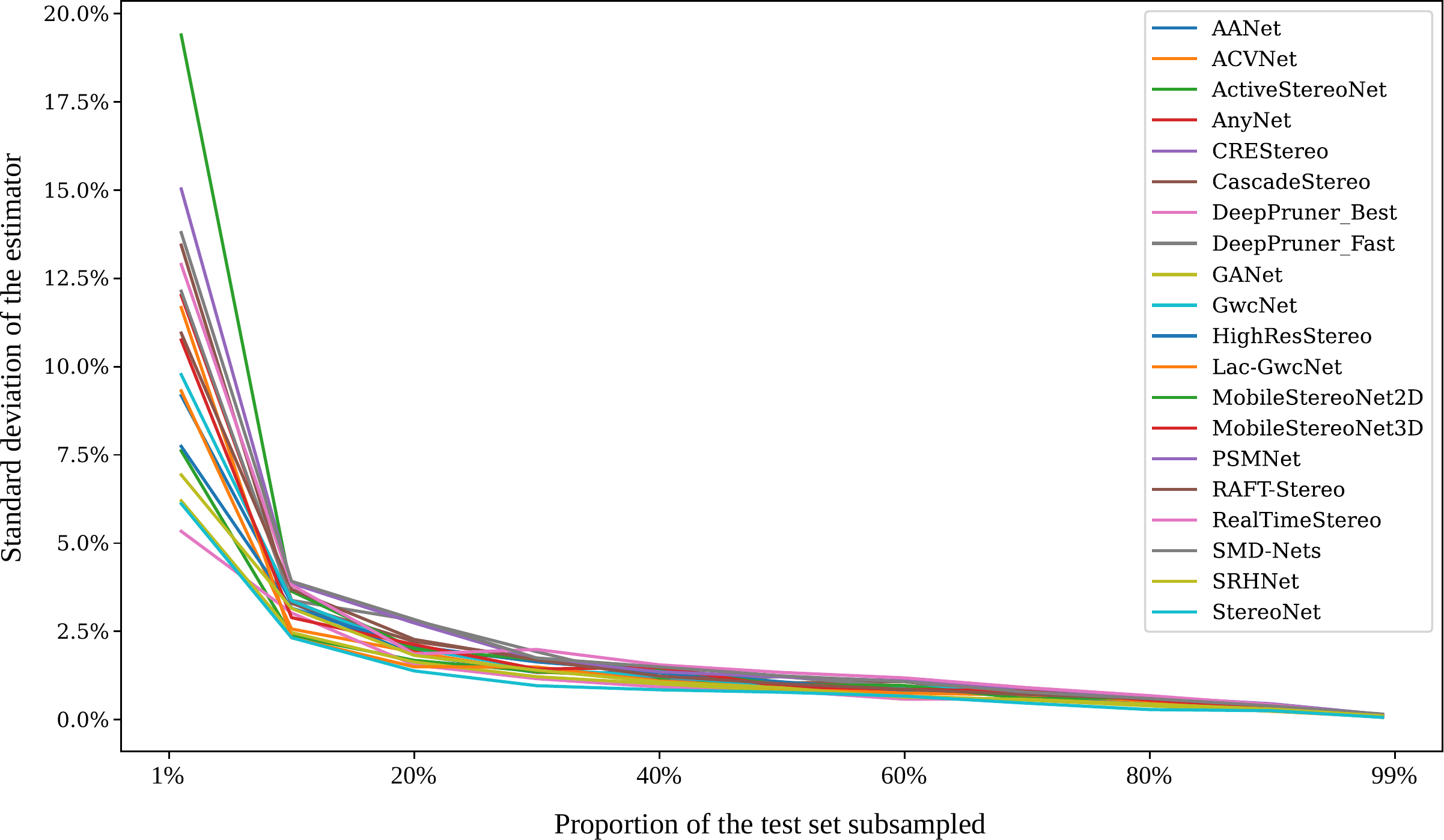}
        \caption{$BAD_2$}
        \label{fig:boostrap_variance:BAD2}
    \end{subfigure}
    
    \caption{Standard deviation of the $MAE$ and $BAD_2$ metrics as a function of the proportion of images subsampled at random in the test set (obtained with the active version of the test set).}
    \label{fig:boostrap_variance}
\end{figure}

The amplitude of $R(M_{\text{pas}}, M_{\text{act}})$ is much larger than the amplitude of $\mathbb{E}[R(M, M_{50\%})]$ when $M$ is either $MAE$ or $BAD_2$. As such, our dataset can be seen as sufficiently large to enable the evaluation of the generalization ability, from passive to active stereo, of deep learning models.

To determine whether the absolute values of the performance metrics are reliable, we analysed the standard deviation $\sqrt{Var[M_{n\%}]}$ of the estimator $M_{n\%}$, where n is a variable parameter. We want to check whether or not $\sqrt{Var[M_{n\%}]}$ gets negligible compared to the metric $M$ and if it tends towards 0 as n grows. We thus computed the value of $\sqrt{Var[M_{n\%}]}$ for $M$ being both $MAE$ and $BAD_2$ and $n$ equal to $1\%$, $10\%$, $20\%$, $30\%$, $40\%$, $50\%$, $60\%$, $70\%$, $80\%$, $90\%$ and $99\%$. The results, reported in Figure~\ref{fig:boostrap_variance}, show clearly that the variance of the estimator stabilizes already towards 0 when $n = 50\%$ well below the variance of the $n = 1\%$ estimate. This shows that the dataset is large enough to get stable estimates of the performances of the different methods on the metrics used for evaluation.

Another point to consider is whether or not the results remains coherent when the test set is re-sampled. If the errors of all models are positively correlated, then selecting a test set that is designed to be more or less challenging should impact all methods in a similar fashion. It turns out that the metrics for each image, and consequently the differences between these metrics, are highly correlated (see \eg Table~\ref{tab:corrcoeffs}, which reports the Pearson correlation coefficients of the difference between the passive and active $MAE$ scores). A notable exception is CREStereo, which, while still clearly positively correlated with all methods, is significantly less correlated with each method than the other methods are correlated among themselves. This is probably due to the high performance of CREStereo on fine details, that other methods struggle to reconstruct. Still, Table~\ref{tab:corrcoeffs} shows that resampling the test set would affect the performances of all methods in a similar fashion.

\begin{table}[]
    \centering
    \caption{Pearson correlation coefficients of the differences between passive and active $MAE$ scores on all images in $T$ for all methods}
    \label{tab:corrcoeffs}
    \vspace{5pt}
    \resizebox{\columnwidth}{!}{
\begin{tabular}{lrrrrrrrrrrrrrrrrrrrr}
\toprule
Method &  \parbox{2mm}{\rotatebox[origin=c]{90}{AANet}} &  \parbox{2mm}{\rotatebox[origin=c]{90}{ACVNet}} &  \parbox{2mm}{\rotatebox[origin=c]{90}{ActiveStereoNet}} &  \parbox{2mm}{\rotatebox[origin=c]{90}{AnyNet}} &  \parbox{2mm}{\rotatebox[origin=c]{90}{CREStereo}} &  \parbox{2mm}{\rotatebox[origin=c]{90}{CascadeStereo}} &  \parbox{2mm}{\rotatebox[origin=c]{90}{DeepPruner\_Best}} &  \parbox{2mm}{\rotatebox[origin=c]{90}{DeepPruner\_Fast}} &  \parbox{2mm}{\rotatebox[origin=c]{90}{GANet}} &  \parbox{2mm}{\rotatebox[origin=c]{90}{GwcNet}} &  \parbox{2mm}{\rotatebox[origin=c]{90}{HighResStereo}} &  \parbox{2mm}{\rotatebox[origin=c]{90}{Lac-GwcNet}} &  \parbox{2mm}{\rotatebox[origin=c]{90}{MobileStereoNet2D}} &  \parbox{2mm}{\rotatebox[origin=c]{90}{MobileStereoNet3D}} &  \parbox{2mm}{\rotatebox[origin=c]{90}{PSMNet}} &  \parbox{2mm}{\rotatebox[origin=c]{90}{RAFT-Stereo}} &  \parbox{2mm}{\rotatebox[origin=c]{90}{RealTimeStereo}} &  \parbox{2mm}{\rotatebox[origin=c]{90}{SMD-Nets}} &  \parbox{2mm}{\rotatebox[origin=c]{90}{SRHNet}} &  \parbox{2mm}{\rotatebox[origin=c]{90}{StereoNet}} \\
Method            &        &         &                  &         &            &                &                  &                  &        &         &                &             &                    &                    &         &              &                 &           &         &            \\
\midrule
AANet             &   100\% &     70\% &              82\% &     74\% &        42\% &            74\% &              95\% &              82\% &    86\% &     92\% &            96\% &         96\% &                96\% &                78\% &     97\% &          93\% &             66\% &       81\% &     92\% &        83\% \\
ACVNet            &    70\% &    100\% &              74\% &     84\% &        42\% &            92\% &              78\% &              91\% &    90\% &     80\% &            72\% &         76\% &                75\% &                83\% &     73\% &          66\% &             63\% &       82\% &     73\% &        67\% \\
ActiveStereoNet   &    82\% &     74\% &             100\% &     77\% &        50\% &            75\% &              81\% &              80\% &    82\% &     83\% &            80\% &         83\% &                87\% &                76\% &     82\% &          72\% &             77\% &       86\% &     80\% &        76\% \\
AnyNet            &    74\% &     84\% &              77\% &    100\% &        38\% &            91\% &              77\% &              91\% &    90\% &     82\% &            74\% &         79\% &                81\% &                87\% &     77\% &          60\% &             75\% &       85\% &     76\% &        75\% \\
CREStereo         &    42\% &     42\% &              50\% &     38\% &       100\% &            36\% &              44\% &              38\% &    39\% &     52\% &            47\% &         47\% &                44\% &                33\% &     44\% &          40\% &             58\% &       40\% &     47\% &        49\% \\
CascadeStereo     &    74\% &     92\% &              75\% &     91\% &        36\% &           100\% &              80\% &              92\% &    91\% &     81\% &            74\% &         81\% &                80\% &                82\% &     75\% &          66\% &             64\% &       85\% &     76\% &        77\% \\
DeepPruner\_Best   &    95\% &     78\% &              81\% &     77\% &        44\% &            80\% &             100\% &              87\% &    89\% &     95\% &            97\% &         99\% &                96\% &                80\% &     94\% &          91\% &             64\% &       85\% &     92\% &        85\% \\
DeepPruner\_Fast   &    82\% &     91\% &              80\% &     91\% &        38\% &            92\% &              87\% &             100\% &    95\% &     87\% &            84\% &         87\% &                87\% &                94\% &     84\% &          76\% &             66\% &       90\% &     78\% &        75\% \\
GANet             &    86\% &     90\% &              82\% &     90\% &        39\% &            91\% &              89\% &              95\% &   100\% &     91\% &            86\% &         89\% &                91\% &                89\% &     89\% &          81\% &             62\% &       88\% &     85\% &        78\% \\
GwcNet            &    92\% &     80\% &              83\% &     82\% &        52\% &            81\% &              95\% &              87\% &    91\% &    100\% &            94\% &         96\% &                95\% &                82\% &     94\% &          84\% &             70\% &       88\% &     91\% &        84\% \\
HighResStereo     &    96\% &     72\% &              80\% &     74\% &        47\% &            74\% &              97\% &              84\% &    86\% &     94\% &           100\% &         97\% &                95\% &                78\% &     96\% &          93\% &             64\% &       82\% &     90\% &        82\% \\
Lac-GwcNet        &    96\% &     76\% &              83\% &     79\% &        47\% &            81\% &              99\% &              87\% &    89\% &     96\% &            97\% &        100\% &                97\% &                82\% &     95\% &          89\% &             67\% &       86\% &     93\% &        87\% \\
MobileStereoNet2D &    96\% &     75\% &              87\% &     81\% &        44\% &            80\% &              96\% &              87\% &    91\% &     95\% &            95\% &         97\% &               100\% &                83\% &     96\% &          89\% &             67\% &       90\% &     90\% &        85\% \\
MobileStereoNet3D &    78\% &     83\% &              76\% &     87\% &        33\% &            82\% &              80\% &              94\% &    89\% &     82\% &            78\% &         82\% &                83\% &               100\% &     81\% &          72\% &             64\% &       88\% &     74\% &        69\% \\
PSMNet            &    97\% &     73\% &              82\% &     77\% &        44\% &            75\% &              94\% &              84\% &    89\% &     94\% &            96\% &         95\% &                96\% &                81\% &    100\% &          92\% &             66\% &       82\% &     91\% &        79\% \\
RAFT-Stereo       &    93\% &     66\% &              72\% &     60\% &        40\% &            66\% &              91\% &              76\% &    81\% &     84\% &            93\% &         89\% &                89\% &                72\% &     92\% &         100\% &             50\% &       73\% &     82\% &        73\% \\
RealTimeStereo    &    66\% &     63\% &              77\% &     75\% &        58\% &            64\% &              64\% &              66\% &    62\% &     70\% &            64\% &         67\% &                67\% &                64\% &     66\% &          50\% &            100\% &       71\% &     69\% &        62\% \\
SMD-Nets          &    81\% &     82\% &              86\% &     85\% &        40\% &            85\% &              85\% &              90\% &    88\% &     88\% &            82\% &         86\% &                90\% &                88\% &     82\% &          73\% &             71\% &      100\% &     81\% &        80\% \\
SRHNet            &    92\% &     73\% &              80\% &     76\% &        47\% &            76\% &              92\% &              78\% &    85\% &     91\% &            90\% &         93\% &                90\% &                74\% &     91\% &          82\% &             69\% &       81\% &    100\% &        83\% \\
StereoNet         &    83\% &     67\% &              76\% &     75\% &        49\% &            77\% &              85\% &              75\% &    78\% &     84\% &            82\% &         87\% &                85\% &                69\% &     79\% &          73\% &             62\% &       80\% &     83\% &       100\% \\
\bottomrule
\end{tabular}

}
\end{table}

\section{The question of color management}
\label{sec:colormanagement}

\begin{figure}[b]
    \centering
   
    \begin{subfig}{0.47\textwidth}
        \centering
       \includegraphics[width=\textwidth]{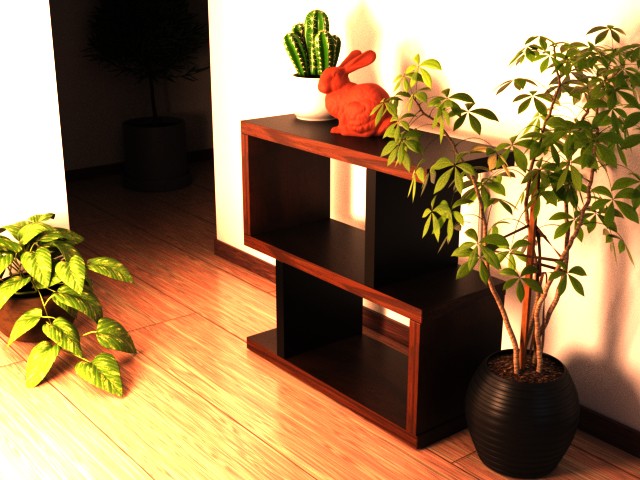}
        \caption{}
       \label{fig:colormanagement:scenereferred}
    \end{subfig}
    \hfill
    \begin{subfig}{0.47\textwidth}
        \centering
        \includegraphics[width=\textwidth]{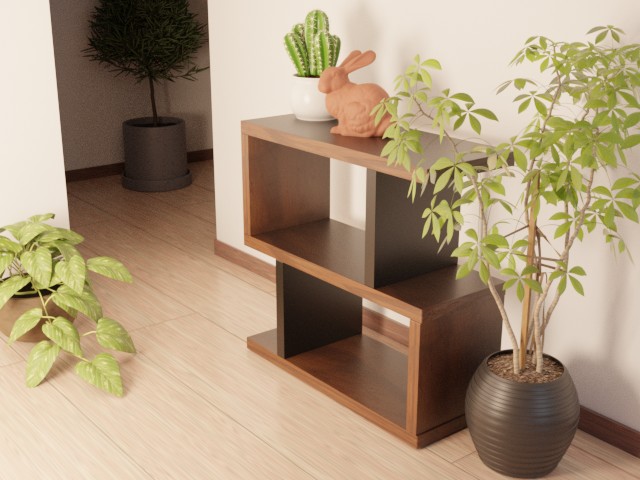}
        \caption{}
        \label{fig:colormanagement:displayreferred}
    \end{subfig}
   
    \caption{The same image in (\subref{fig:colormanagement:scenereferred}) scene referred colors (shown as if display referred) and (\subref{fig:colormanagement:displayreferred}) in display referred colors.}
    \label{fig:colormanagement}
\end{figure}

An important point which is often overlooked in computer vision datasets is the question of color management. Color management, in computer science, generally refers to transformations across color representation models (\eg RGB, HSV, HSL, and XYZ). It can also refer to the collection of processes aiming at preserving the appearance of images across different devices (\eg linear RGB \vs    perceptual RGB) \cite{10.1145/2343483.2343492}.

Modern pipelines for computer-generated imagery, including the one used for our simulations, are constructed around a scene-referred color space; see Fig.~\ref{fig:colormanagement:scenereferred}. In other words,  the numerical value that encodes a color is directly proportional to the light intensity in the scene. However, the image formats used for exchange and storage generally use a display-referred color space; see Fig.~\ref{fig:colormanagement:displayreferred}). In other words, the numerical value encoding a color is proportional to the command that will be sent to the display device, either a screen or a projector. As the dynamic range available on such display devices is far lower than in the acquired scenes, images need to undergo a process called tone mapping, which applies a non-linear transformation to map the scene-referred color space to the display-referred color space \cite{10.1145/2343483.2343492}.

We chose to keep a scene-referred color space for our dataset as it preserves more information. The display-referred images are also available with the published dataset. For the evaluation, we recommend choosing the color space that matches the intended application of the method. A method aimed at processing the raw signal from a camera has to be evaluated on the scene-referred images while methods taking usual image files as input need to be tested on display-referred images. When benchmarking existing methods on our dataset, we evaluated both color spaces and kept the best performing one for each of the methods we tested.

We rely on OpenColorIO \cite{10.1145/3450508.3464600} to perform the different color transformations between scene and display-refereed colors. OpenColorIO is an open source library available online \cite{ocio}. The library requires configuration files encoding the different color transformations. We used the configuration files provided with the open source software Blender \cite{BlenderSoftware} (on version 2.93, but the files provided with the newer Blender 3.0 and 3.1 are very similar and are thus usable as well). Our post-processing code is also publicly available along with the dataset.

\section{Creating your own simulated images}
\label{sec:simulator}

\begin{figure}[tb]
    \centering
    \includegraphics[width=\textwidth]{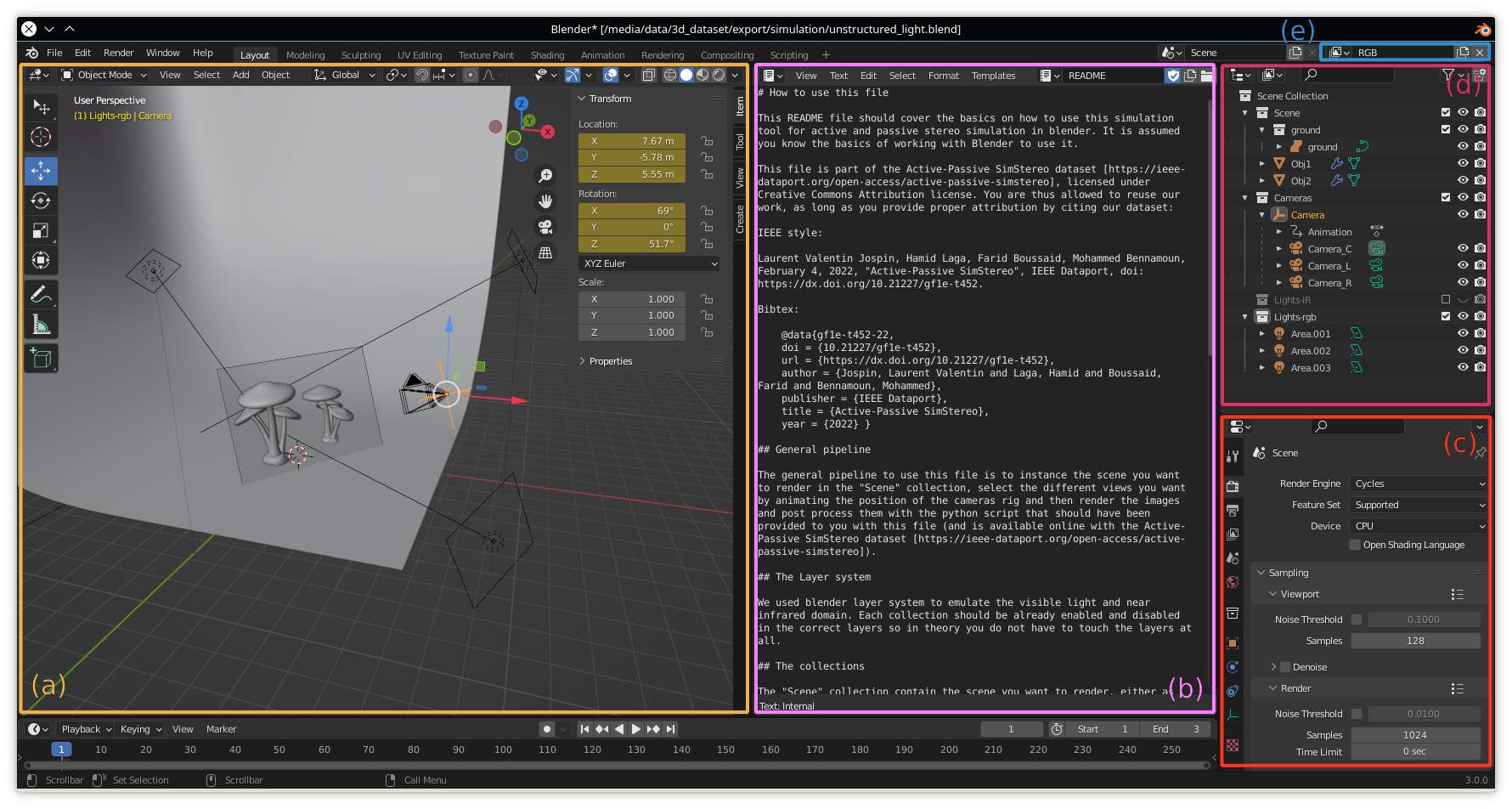}
    \caption{The interface of our simulation setup in Blender.}
    \label{fig:simulator}
\end{figure}

We provided our base simulation Blender file on IEEE dataport to let interested readers create their own simulated active-passive stereo images. This section  introduces the pipeline in the file. Here, we assume a certain level of familiarity with the Blender software.

The assets need to be imported into the main 3D workspace (Fig.~\ref{fig:simulator}a) where the scene can be composed. Next to the 3D workspace, you can find the text editor with an internal README file opened (Fig.~\ref{fig:simulator}b), presenting  the main pipeline. 

There are a few parameters that need to be checked before rendering in the properties editor (Fig.~\ref{fig:simulator}c). Make sure you:
\begin{enumerate}
    \item are rendering with the Cycles rendering engine,
    \item have multi-view rendering enabled and set the rendering folder and filename template to sensible values,
    \item render to Multilayer EXR images, and tick to have a single file with multiple views instead of one file per view.
\end{enumerate}

\noindent Those parameters are pre-selected for you, but future versions of Blender might replace some of the default options. You might also want to tweak your rendering setting. The generated active pattern can interfere with the denoising systems in Cycles, so make sure to turn them off.

Rendering with a Ray-tracing engine can be quite slow. Table~\ref{tab:render_timing} reports average rendering times measured on our hardware configuration (NVIDIA GeForce RTX 2080 Ti GPU and Intel i9-10900X CPU) for a subset of frames. Rendering times may vary depending on the complexity of the scene at hand. The latest version of the Cycles rendering engine, Cycles X, support either optix or cuda as back-end on the GPU. We recommend you use GPU rendering.  If your GPU supports it, we recommend you use optix.

\begin{table}[tb]
    \centering
    
    \caption{Typical rendering time for $640\times480$ simulated images on an NVIDIA GeForce RTX 2080 Ti GPU and Intel i9-10900X CPU}
    \label{tab:render_timing}
    \vspace{5pt}
    \resizebox{\columnwidth}{!}{
    \begin{tabular}{lrrrcrrr}
\toprule
{} & Cycles X (optix) & Cycles X (cuda) & Cycles X (cpu) \\
\midrule
Passive frame & 0m50s +/- 09s & 2m48s +/- 0m26s & 9m37s +/- 0m49s \\
Active frame & 1m19s +/- 19s & 5m25s +/- 0m51s & 16m17s +/- 1m17s \\
Total (2x passive + 2x active) & 4m18s +/- 56s & 16m26s +/- 2m34s & 51m48s +/- 4m12s \\
\bottomrule

\end{tabular}
    }
\end{table}

When you import your assets to create your scene, make sure to place them in the Scene collection in the outliner (Fig.~\ref{fig:simulator}d) as other collections might be turned on or off in certain rendering layers. If you want to edit your light for passive stereo, make sure to jump to the ``Simulated NIR" view layer (Fig.~\ref{fig:simulator}e). If you want to go back to the visible light layer, return to the ``RGB" view layer.

If you want to move or animate the camera, move the ``Camera" empty object and not the three children ``Camera\_L", ``Camera\_C" and ``Camera\_R". This will keep the relative positions of the different cameras consistent.

Once you are happy with your composition, you can render a single image or an image sequence with a moving cameras. You then just need to post-process them with the ``post\_process\_exr.py" Python script provided aside the simulation file.

We will also release additional data packs in the future, as we simulate more images for additional applications. At the time of writing, a first data pack with 7500 additional images from 5 scenes has been released.

\section{Long-term storage}
\label{sec:storage}

We are using IEEE dataport \cite{ieeedataport} for the long-term storage of the dataset. To alleviate the issue of the required subscription to access the data, the \datasetname{} dataset \cite{our_dataset} has been published in open access. The only requirement to download the files is to get an IEEE account, which is free.

This should ensure that the \datasetname{} dataset and the associated tools will stay available in the foreseeable future.

\bibliographystyle{plainnat}
\bibliography{bibliography}